\documentclass[10pt,twocolumn,letterpaper]{article}

\usepackage{cvpr}
\usepackage{times}
\usepackage{epsfig}
\usepackage{graphicx}
\usepackage{amsmath}
\usepackage{amssymb}

\usepackage{bm}
\usepackage{amsmath}
\usepackage{amsfonts}
\usepackage{booktabs}
\usepackage{subcaption}
\usepackage{threeparttable}
\usepackage{url}

\usepackage[breaklinks=true,bookmarks=false]{hyperref}

\cvprfinalcopy 

\def\cvprPaperID{****} 

\begin{document}

\title{HoMM: Higher-order Moment Matching for Unsupervised Domain Adaptation}

\author{Chao Chen\textsuperscript{\rm 1,\rm 2}\thanks{This work was done as a research intern in Alibaba Group. This work is supported by the opening foundation of the State Key Laboratory (No. 2014KF06) and CSC Scholarship.}, Zhihang Fu\textsuperscript{\rm 2}, Zhihong Chen\textsuperscript{\rm 1}, Sheng Jin\textsuperscript{\rm 2},\\
Zhaowei Cheng\textsuperscript{\rm 1}, Xinyu Jin\textsuperscript{\rm 1}, Xian-Sheng Hua\textsuperscript{\rm 2}\thanks{This is the corresponding author.}  \\
\textsuperscript{\rm 1} Zhejiang University, \quad \textsuperscript{\rm 2} Alibaba DAMO Academy, Alibaba Group\\
\tt{chench@zju.edu.cn}, \tt{xiansheng.hxs@alibaba-inc.com}
}


\maketitle

\begin{abstract}
Minimizing the discrepancy of feature distributions between different domains is one of the most  promising directions in unsupervised domain adaptation. From the perspective of distribution matching, most existing discrepancy-based methods are designed to match the second-order or lower statistics, which however, have limited expression of statistical characteristic for non-Gaussian distributions. In this work, we explore the benefits of using higher-order statistics (mainly refer to third-order and fourth-order statistics) for domain matching. We propose a Higher-order Moment Matching (HoMM) method, and further extend the HoMM into reproducing kernel Hilbert spaces (RKHS). In particular, our proposed HoMM can perform arbitrary-order moment tensor matching, we show that the first-order HoMM is equivalent to Maximum Mean Discrepancy (MMD) and the second-order HoMM is equivalent to Correlation Alignment (CORAL). Moreover, the third-order and the fourth-order moment tensor matching are expected to perform comprehensive domain alignment as higher-order statistics can approximate more complex, non-Gaussian distributions. Besides, we also exploit the pseudo-labeled target samples to learn discriminative representations in the target domain, which further improves the transfer performance. Extensive experiments are conducted, showing that our proposed HoMM consistently outperforms the existing moment matching methods by a large margin. Codes are available at \url{https://github.com/chenchao666/HoMM-Master}
\end{abstract}

\section{Introduction}
Convolutional neural networks (CNNs) have shown promising results on supervised learning tasks. However, the performance of a learned model always degrades severely when dealing with data from the other domains. Considering that constantly annotating massive samples from new domains is expensive and impractical, unsupervised domain adaptation (UDA), therefore, has emerged as a new learning framework to address this problem \cite{csurka2017domain}. UDA aims to utilize full-labeled samples in source domain to annotate the completely-unlabeled target domain samples. Thanks to deep CNNs, recent advances in UDA show satisfactory performance in several computer vision tasks \cite{hoffman2018cycada}. Among them, most methods bridge the source and target domain by learning domain-invariant features. These dominant methods can be further divided into two categories: (1) Learning domain-invariant features by minimizing the discrepancy between feature distributions \cite{long2015learning,sun2016deep,long2017deep,zellinger2017central,chen2019towards}. (2) Encouraging domain confusion by a domain adversarial objectives whereby a discriminator (domain classifier) is trained to distinguish between the source and target representations. \cite{ganin2016domain,tzeng2017adversarial,shu2018dirt,hoffman2018cycada}.

\begin{figure}[t!]
\centering
\includegraphics[width=1.0\linewidth]{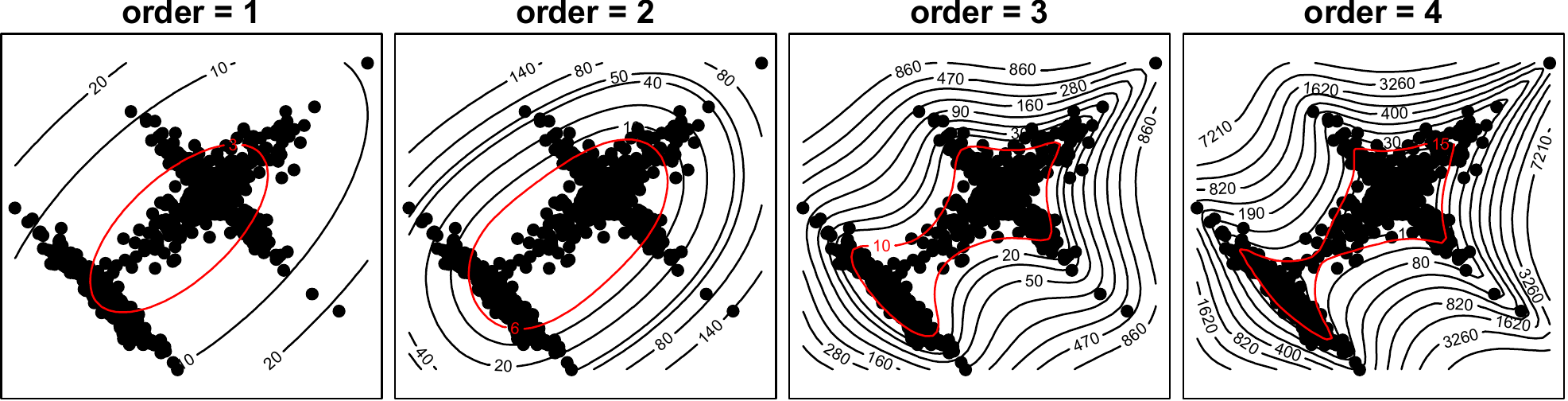}
\caption{The metrics of using higher-order moment tensor for domain alignment. 300 points in $\mathbb{R}^2$ and the level set of the moment tensor with different order. As observed, using higher-order moment tenser captures the shape of the cloud of samples more accurately.}
\label{fig0}
\end{figure}

From the perspective of moment matching, most of the existing discrepancy-based methods in UDA are based on Maximum Mean Discrepancy (MMD) \cite{long2017deep} or Correlation Alignment (CORAL) \cite{sun2016deep}, which are designed to match the first-order (Mean) and second-order (Covariance) statistics of different distributions. However, for the real world applications (such as image recognition), the deep features are always a complex, non-Gaussian distribution \cite{jia2011heavy,xu2016blind}, which can not be completely characterized by its first-order or second-order statistics. Therefore, aligning the second-order or lower statistics only guarantees coarse-grained alignment of two distributions. To address this limitation, we propose to perform domain alignment by matching the higher-order moment tensor (mainly refer to third-order and fourth-order moment tensor), which contain more discriminative information and can better represent the feature distribution \cite{pauwels2016sorting,gou2017mom}. Inspired by \cite{pauwels2016sorting}, Fig.\ref{fig0} illustrates the metrics of higher-order moment tensor, where we plot a cloud of points (consists of three different Gaussians) and the level sets of moment tensor with different order.  As observed, the higher-order moment tensor characterizes the distribution more accurately.

Our contribution can be concluded as: (1) We propose a Higher-order Moment Matching (HoMM) method to minimize the domain discrepancy, which is expected to perform fine-grained domain alignment. The HoMM integrates the MMD and CORAL into a unified framework and generalizes the first-order and second-order moment matching to higher-order moment tensor matching. Without bells and whistles, the third- and fourth-order moment matching outperform all existing discrepancy-based methods by a large margin. The source code of the HoMM is released. (2) Due to lack of labels in the target domain, we propose to learn discriminative clusters in the target domain by assigning the pseudo-labels for the reliable target samples, which also improves the transfer performance.

\section{Related Work}
\textbf{Learning Domain-Invariant Features} To minimize the domain discrepancy and learn domain-invariant features, various distribution discrepancy metrics have been introduced. The representative ones include Maximal Mean Discrepancy (MMD) \cite{gretton2012kernel,tzeng2014deep,long2015learning,long2017deep},  Correlation Alignment \cite{sun2016deep,morerio2018,chen2019joint,chen2019selective} and Wasserstein distance \cite{lee2019sliced,chendeep}. MMD was first introduced for the two-sample tests problem \cite{gretton2012kernel}, and is currently the most widely used metric to measure the distance between two feature distributions. Specifically, Long \etal proposed DAN \cite {long2015learning} and JAN \cite{long2017deep} which perform domain matching via multi-kernel MMD or a joint MMD criteria in multiple domain-specific layers across domains. Sun \etal proposed the correlation alignment (CORAL) \cite{sun2016return,sun2016deep} to align the second order statistics of the source and target distributions. Some recent work also extended the CORAL into reproducing kernel Hilbert spaces (RKHS) \cite{zhang2018aligning} or deployed alignment along geodesics by considering the log-Euclidean distance \cite{morerio2018}. Interestingly, \cite{li2017demystifying} theoretically demonstrated that matching the second order statistics is equivalent to minimizing MMD with the second order polynomial kernel. Besides, the approach most relevant to our proposal is the Central Moment Discrepancy (CMD) \cite{zellinger2017central}, which matches the higher order central moments of probability distributions by means of order-wise moment differences. Both CMD and our HoMM propose to match the higher-order statistics for domain alignment. The CMD matches the higher-order central moment while our HoMM matches the higher-order cumulant tensor. Another fruitful line of work tries to learn the domain-invariant features through adversarial training \cite{ganin2016domain,tzeng2017adversarial}. These efforts encourage domain confusion by a domain adversarial objective whereby a discriminator (domain classifier) is trained to distinguish between the source and target representations. Also, recent work performing pixel-level adaptation by image-to-image transformation \cite{murez2018image,hoffman2018cycada} has achieved satisfactory performance and obtained much attention. In this work, we propose a higher-order moment tensor matching approach to minimize the domain discrepancy, which shows great superiority over existing discrepancy-based methods.

\noindent\textbf{Higher-order Statistics} The statistics higher than first-order has been success fully used in many classical and deep learning methods \cite{de2007fourth,koniusz2016higher,pauwels2016sorting,li2017second,gou2017mom}. Especially in the field of fine-grained image/video recognition \cite{lin2015bilinear,cui2017kernel}, second-order statistics such as Covariance and Gaussian descriptors, have demonstrated better performance than descriptors exploiting zeroth- or first-order statistics \cite{li2017second,lin2015bilinear,wang2017g2denet}. However, using second-order or lower statistical information might not be enough when the feature distribution is non-Gaussian \cite{gou2017mom}. Therefore, the higher-order (greater than two) statistics have been explored in many signal processing problems \cite{mansour1995fourth,jakubowski2002higher,de2007fourth,gou2017mom}. In the field of Blind Source Separation (BSS) \cite{de2007fourth,mansour1995fourth}, for example, the fourth-order statistics are widely used to identify different signals from mixtures. Gou \etal utilizes the third-order statistics for person ReID \cite{gou2017mom}, Xu \etal exploits the third-order cumulant for blind image quality assessment \cite{xu2016blind}. In \cite{jakubowski2002higher,koniusz2016higher}, the authors exploit higher-order statistics for image recognition and detection. Matching the second order statistics can not always ensure two distributions inseparable, just as using the second order statistics can not identifies different signals from underdetermined mixtures \cite{de2007fourth}. That's why we explore higher-order moment tensor for domain alignment.

\noindent\textbf{Discriminative Clustering} Discriminative clustering is a critical task in the unsupervised and semi-supervised learning paradigms \cite{grandvalet2005semi,lee2013pseudo,xie2016unsupervised,caron2018deep}. Due to the paucity of labels in the target domain, how to obtain the discriminative representations in the target domain is of great importance for the UDA tasks. Therefore, a large body of work pays attention to learn the discriminative clusters in the target domain via entropy minimization \cite{grandvalet2005semi,morerio2018,shu2018dirt}, pseudo label \cite{lee2013pseudo,saito2017asymmetric,xie2018learning} or distance-based metrics \cite{chen2019joint,kang2019contrastive}. Specifically, Saito \etal \cite{saito2017asymmetric} assign pseudo-labels to the reliable unlabeled samples to learn discriminative representations for the target domain. Shu \etal \cite{shu2018dirt} consider the cluster assumption and minimize the conditional entropy to ensure the decision boundaries not cross high-density data regions. MCD \cite{saito2018maximum} also considers to align distributions of source and target by utilizing the task-specific decision boundaries. Besides, JDDA \cite{chen2019joint} and CAN \cite{kang2019contrastive} propose to model the intra-class domain discrepancy and the inter-class domain discrepancy to learn more discriminative features.

\section{Method}
\begin{figure}[t!]
\centering
\includegraphics[width=0.8\linewidth]{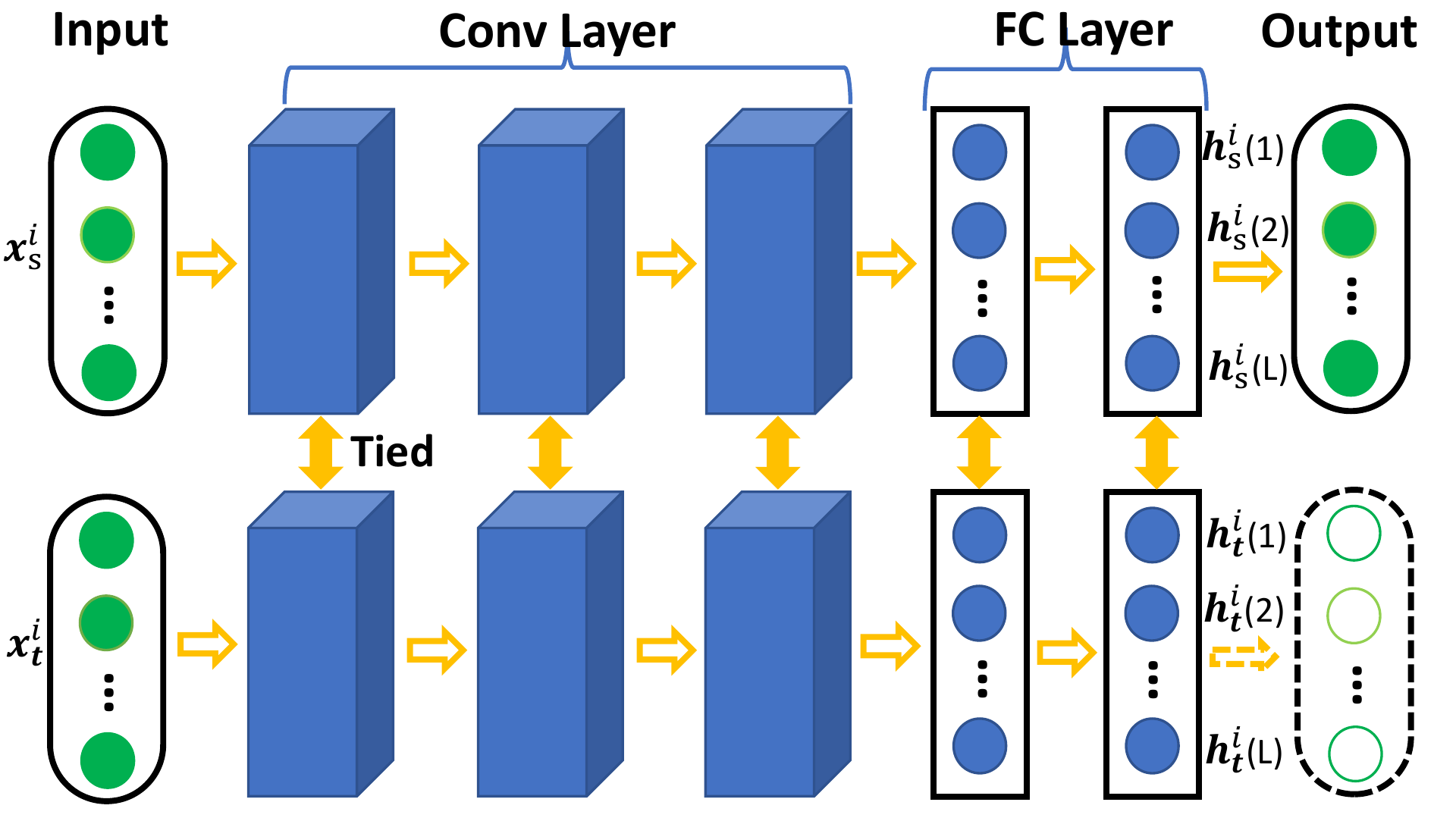}
\caption{Two-stream CNNs with shared parameters are adopted for unsupervised deep domain adaptation. The first stream operates the source data and the second stream operates the target data. The last FC layer (the input of the output layer) is used as the adapted layer.}
\label{fig1}
\end{figure}
In this work, we consider the unsupervised domain adaptation problem. Let $\mathcal{D}_s=\{\bm{x}_s^i,y_s^i\}_{i=1}^{n_s}$ denotes the source domain with $n_s$ labeled samples and $\mathcal{D}_t=\{\bm{x}_t^i\}_{i=1}^{n_t}$ denotes the target domain with $n_t$ unlabeled samples. Given $\mathcal{D}_s\cup\mathcal{D}_t$, we aim to train a cross-domain CNN classifier $f_{\bm\theta}(\bm{x})$ which can minimize the target risks $\epsilon_t=\mathbb{E}_{\bm{x}\in\mathcal{D}_t}[f_{\bm\theta}(\bm{x})\neq \bm{y}_t]$. Here $f_{\bm\theta}(\bm{x})$ denotes the outputs of the deep neural networks, $\bm\theta$ denotes the model parameter to be learned. Following \cite{long2017deep,chen2019joint}, we adopt the two-stream CNNs architecture for unsupervised deep domain adaptation.  As shown in Fig. \ref{fig1}, the two streams share the same parameters (tied weights), operating the source and target domain samples respectively. And we perform the domain alignment in the last full-connected (FC) layer \cite{sun2016deep,chen2019joint}. According to the theory proposed by Ben-David \etal \cite{ben2010theory}, a basic domain adaptation model should, at least, involve the source domain loss and the domain discrepancy loss, i.e.,
\begin{equation}\label{eq1}
\mathcal{L}(\bm\theta|\mathbf{X}_s,\mathbf{Y}_s,\mathbf{X}_t)=\mathcal{L}_s+\lambda_{d}\mathcal{L}_{d}
\end{equation}
\begin{equation}\label{eq2}
\mathcal{L}_s=\dfrac{1}{n_s}\sum\limits_{i=1}^{n_s}J(f_{\bm\theta}(\bm{x}_i^s), \bm{y}_i^s)
\end{equation}
where $\mathcal{L}_s$ represents the classification loss in the source domain, $J(\cdot,\cdot)$ represents the cross-entropy loss function. $\mathcal{L}_{d}$ represents the domain discrepancy loss and $\lambda_{d}$ is the trade-off parameter. As aforementioned, most of existing discrepancy-based methods are designed to minimize distance of the second-order or lower statistics between different domains. In this work, we propose a higher-order moment matching method, which matches the higher-order statistics of different domains.
\subsection{Higher-order Moment Matching}
To perform fine-grained domain alignment, we propose a higher-order moment matching as
\begin{equation}\label{eq3}
\mathcal{L}_d=\frac{1}{L^p}\Vert\frac{1}{n_s}\sum\limits_{i=1}^{n_s}\phi_{\bm\theta}(\bm{x}_s^i)^{\otimes p}-\frac{1}{n_t}\sum\limits_{i=1}^{n_t}\phi_{\bm\theta}(\bm{x}_t^i)^{\otimes p}\Vert_F^2
\end{equation}
where $n_s=n_t=b$ ($b$ is the batch size) during the training process. $\phi_{\bm\theta}(\bm{x})$ denotes the activation outputs of the adapted layer. As illustrated in Fig. \ref{fig1}, $\bm{h}^i=\phi_{\bm\theta}(\bm{x}^i)=[\bm{h}^i(1),\bm{h}^i(2),\cdots,\bm{h}^i(L)]\in\mathbb{R}^L$ denotes the activation outputs of the $i$-th sample, $L$ is the number of hidden neurons in the adapted layer. Here, $\bm{u}^{\otimes p}$ denotes the $p$-level tensor power of the vector $\bm{u}\in\mathbb{R}^c$. That is
\begin{equation}\label{eq4}
\bm{u}^{\otimes p}=\underbrace{\bm{u}\otimes\bm{u}\cdots\otimes\bm{u}}_{\text{p times}}\in\mathbb{R}^{c^p}
\end{equation}
where $\otimes$ denotes the outer product (or tensor product). We have $\bm{u}^{\otimes 0}=1$, $\bm{u}^{\otimes 1}=\bm{u}$ and $\bm{u}^{\otimes 2}=\bm{u}\otimes\bm{u}$. The 2-level tensor product $\bm{u}^{\otimes 2}\in\mathbb{R}^{c^2}$ defined as
\begin{equation}\label{eq4}
\bm{u}^{\otimes 2}=\bm{u}^T\bm{u}=\left[\begin{array}{cccc}
u_1u_1 & u_1u_2 & \cdots & u_1u_c\\
u_2u_1 & u_2u_2 & \cdots & u_2u_c\\
\vdots & \vdots & \ddots & \vdots\\
u_cu_1 & u_cu_2 & \cdots & u_cu_c
\end{array}\right]
\end{equation}
when $p\geq 3$, $\mathbf{T}=\bm{u}^{\otimes p}$ is a $p$-level tensor with $\mathbf{T}[i,j,\cdots,k]=u_iu_j\cdots u_k$.

\begin{figure}[t!]
\centering
\includegraphics[width=1.0\linewidth]{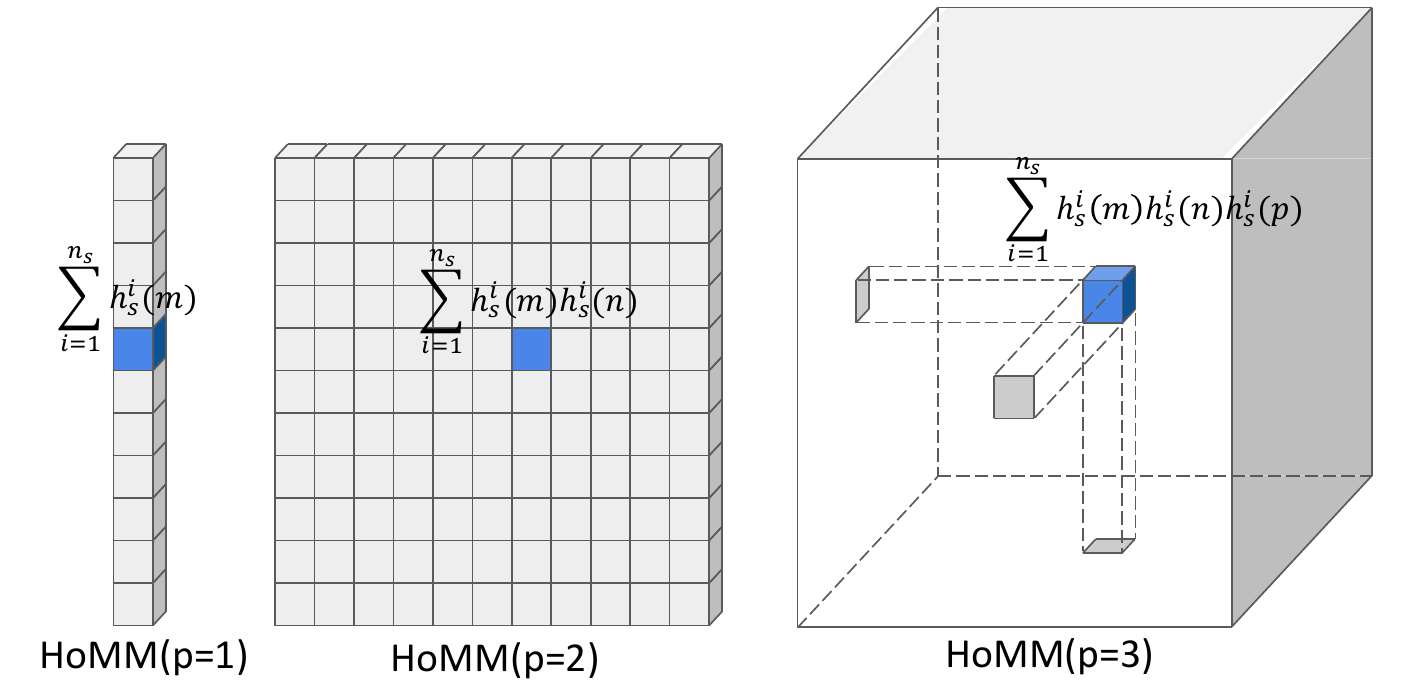}
\caption{An illustration of first-order, second-order and third-order moments in the source domain. HoMM matches the higher-order ($p\geq3$) moment across different domains.}
\label{fig2}
\end{figure}

\textbf{Instantiations}
According to Eq. \eqref{eq3}, when $p=1$, the first-order moment matching is equivalent to the linear MMD \cite{tzeng2014deep}, which is expressed as
\begin{equation}\label{eq6}
\mathcal{L}_d=\frac{1}{L}\Vert\frac{1}{b}\sum\limits_{i=1}^{b}\bm{h}_s^i-\frac{1}{b}\sum\limits_{i=1}^{b}\bm{h}_t^i\Vert_F^2
\end{equation}
When $p=2$, the second-order HoMM is formulated as,
\begin{equation}\label{eq7}
\begin{split}
\mathcal{L}_d=&\frac{1}{L^2}\Vert\frac{1}{b}\sum\limits_{i=1}^{b}\bm{h}_s^i{^T}\bm{h}_s^i-\frac{1}{b}\sum\limits_{i=1}^{b}\bm{h}_t^i{^T}\bm{h}_t^i\Vert_F^2\\
=&\frac{1}{b^2L^2}\Vert\bm{G}(\bm{h}_s)-\bm{G}(\bm{h}_t)\Vert_F^2
\end{split}
\end{equation}
where $\bm{G}(\bm{h})=\bm{H}^T\bm{H}\in\mathbb{R}^{L\times L}$ is the Gram matrix, $\bm{H}=[\bm{h}^1;\bm{h}^2;\cdots,\bm{h}^b]\in\mathbb{R}^{b\times L}$, $b$ is the batch size. Therefore, the second-order HoMM is equivalent to the Gram matrix matching, which is also widely used for cross-domain matching in neural style transfer \cite{gatys2016image,li2017demystifying} and knowledge distillation \cite{yim2017gift}. Li \etal \cite{li2017demystifying} theoretically demonstrate that matching the Gram matrix of feature maps is equivalent to minimize the MMD with the second order polynomial kernel. Besides, when the activation outputs are normalized by subtracting the mean value, the centralized Gram matrix turns into the Covariance matrix. In this respect, the second-order HoMM is also equivalent to CORAL, which matches the Covariance matrix for domain matching \cite{sun2016deep}.

As illustrated in Fig. \ref{fig2}, in addition to the first-order moment matching (e.g. MMD) and the second-order moment matching (e.g. CORAL and Gram matrix matching), our proposed HoMM can also perform higher-order moment tensor matching when $p\geq 3$. Since higher-order statistics can characterize the non-Gaussian distributions better, applying higher-order moment matching is expected to perform fine-grained domain alignment. However, the space complexity of calculating the higher-order tensor $\bm{u}^{\otimes p}$ ($p\geq3$) reaches $\mathcal{O}(L^p)$, which makes the higher-order moment matching infeasible in many real-world applications. Adding bottleneck layers to shrink the length of adaptive layer does not even solve the problem. When $L=128$, for example, the dimension of a third-order tensor still reaches $\mathcal{O}(10^6)$, and the dimension of a fourth-order tensor reaches $\mathcal{O}(10^8)$, which is absolutely computational-unfriendly. To address this problem, we propose two practical techniques to perform the compact tensor matching.

\textbf{Group Moment Matching.} As the space complexity grows exponentially with the number of neurons $L$, one practical approach is to divide the hidden neurons in the adapted layer into $n_g$ groups, with each group $\lfloor L/n_g\rfloor$ neurons. Then we can calculate and match the high-level tensor in each group respectively. That is,
\begin{equation}\label{eq8}
\mathcal{L}_d=\frac{1}{b^2\lfloor L/n_g\rfloor^p}\sum\limits_{k=1}^{n_g}\Vert\sum\limits_{i=1}^{b}\bm{h}_{s,k}^i{^{\otimes p}}-\sum\limits_{i=1}^{b}\bm{h}_{t,k}^i{^{\otimes p}}\Vert_F^2
\end{equation}
where $\bm{h}_{:,k}^i\in\mathbb{R}^{\lfloor L/n_g\rfloor}$ is the activation outputs of $k$-th group. In this way, the space complexity can be reduced from $\mathcal{O}(L^p)$ to $\mathcal{O}(n_g\cdot\lfloor L/n_g\rfloor^p)$. In practice, $\lfloor L/n_g\rfloor\geq25$ need to be satisfied to ensure satisfactory performance.

\textbf{Random Sampling Matching.} The group moment matching can work well when $p=3$ and $p=4$, but it tends to fail when $p\geq5$. Therefore, we also propose a random sampling matching strategy which is able to perform arbitrary-order moment matching. Instead of calculating and matching two high-dimensional tensors, we randomly select $N$ values in the high-level tensor, and only calculate and align these $N$ values in the source and target domains. In this respect, the $p$-order moment matching with random sampling strategy can be formulated as,
\begin{equation}\label{eq9}
\small
\mathcal{L}_d=\frac{1}{b^2N}\sum\limits_{k=1}^{N}[\sum\limits_{i=1}^{b}\prod_{j=rnd[k,1]}^{rnd[k,p]}\bm{h}_s^i(j)-\sum\limits_{i=1}^{b}\prod_{j=rnd[k,1]}^{rnd[k,p]}\bm{h}_t^i(j)]^2
\end{equation}
where $rnd\in\mathbb{R}^{N\times p}$ denotes the randomly generated position index matrix, $rnd[k,j]\in\{1,2,3,\cdots,L\}$. Therefore, $\prod_{j=rnd[k,1]}^{rnd[k,p]}\bm{h}_s^i(j)$ denotes a randomly sampled value in the $p$-level tensor $\bm{h}_{s}^i{^{\otimes p}}$. With the random sampling strategy, we can perform arbitrarily-order moment matching, and the space complexity can be reduced from $\mathcal{O}(L^p)$ to $\mathcal{O}(N)$. In practice, the model can achieve very competitive results even $N=1000$.
\subsection{Higher-order Moment Matching in RKHS}
Similar to the KMMD \cite{long2017deep}, we generalize the higher-order moment matching into reproducing kernel Hilbert spaces (RKHS). That is,
\begin{equation}\label{eq10}
\begin{split}
\mathcal{L}_d=\frac{1}{L^p}\Vert\frac{1}{b}\sum\limits_{i=1}^{b}\psi(\bm{h}_s^i{^{\otimes p}})-\frac{1}{b}\sum\limits_{i=1}^{b}\psi(\bm{h}_t^i{^{\otimes p}})\Vert_F^2
\end{split}
\end{equation}
where $\psi(\bm{h}_s^i{^{\otimes p}})$ denotes the feature representation of $i$-th source sample in RKHS. According to the proposed random sampling strategy, $\bm{h}_s^i{^{\otimes p}}$ and $\bm{h}_t^i{^{\otimes p}}$ can be approximated by two $N$-dimensional vectors $\bm{h}_{sp}^i\in\mathbb{R}^N$ and $\bm{h}_{tp}^i\in\mathbb{R}^N$, where $\bm{h}_{sp}^i(k)=\prod_{j=rnd[k,1]}^{rnd[k,p]}\bm{h}_s^i(j)$, $k=1,\cdots,N$. In this respect, the domain matching loss can be formulated as
\begin{equation}\label{eq11}
\begin{split}
\mathcal{L}_d&=\frac{1}{b^2}\sum\limits_{i=1}^{b}\sum\limits_{j=1}^{b}k(\bm{h}_{sp}^i,\bm{h}_{sp}^j)-\frac{2}{b^2}\sum\limits_{i=1}^{b}\sum\limits_{j=1}^{b}k(\bm{h}_{sp}^i,\bm{h}_{tp}^j) \\
&+\frac{1}{b^2}\sum\limits_{i=1}^{b}\sum\limits_{j=1}^{b}k(\bm{h}_{tp}^i,\bm{h}_{tp}^j)
\end{split}
\end{equation}
where $k(\bm{x},\bm{y})=\exp({-\gamma\Vert \bm{x}-\bm{y}\Vert_2})$ is the RBF kernel function. Particularly, when $p=1$, the kernelized HoMM (KHoMM) is equivalent to the KMMD.

\subsection{Discriminative Clustering}
When the target domain features are well aligned with the source domain features, the unsupervised domain adaptation turns into the semi-supervised classification problem, where the discriminative clustering in the unlabeled data is always encouraged \cite{grandvalet2005semi,xie2016unsupervised}. There have been a lot of work trying to learn the discriminative clusters in the target domain \cite{shu2018dirt,morerio2018}, most of which minimize the conditional entropy to ensure the decision boundaries do not cross high-density data regions,
\begin{equation}\label{eq12}
\mathcal{L}_{ent}=\frac{1}{n_t}\sum_{i=1}^{n_t}\sum_{j=1}^{c}-p_j\log p_j
\end{equation}
where $c$ is the number of classes, $p_j$ is the softmax output of $j$-th node in the output layer. We find that the entropy regularization works well when the target domain has high test accuracy, but it helps little or even downgrades the accuracy when the test accuracy is unsatisfactory. The reason can be drawn that the classifier may be misled as a result of entropy regularization enforcing over-confident probability on some misclassified samples. Instead of clustering in the output layer by minimizing the conditional entropy, we propose to cluster in the shared feature space. First, we pick up highly confident predicted target samples whose predicted probabilities are greater than a given threshold $\eta$, and assign pseudo-labels to these reliable samples. Then, we penalize the distance of each pseudo-labeled sample to its class center. The discriminative clustering loss can be given as
\begin{equation}\label{eq13}
\mathcal{L}_{dc}=\frac{1}{n_t}\sum_{i=1}^{n_t}\Vert\bm{h}_t^i-\bm{c}_{\hat{y}_t^i}\Vert_2^2
\end{equation}
where $\hat{y}_t^i$ is the assigned pseudo-label of $\bm{x}_t^i$, $\bm{c}_{\hat{y}_t^i}\in \mathbb{R}^L$ denotes its estimated class center. As we perform update based on mini-batch, the centers can not be accurately estimated by a small size of samples. Therefore, we update the class center in each iteration via moving average method. That is,
\begin{equation}\label{Eq14}
\mathbf{c}_j^{t+1}=\alpha\mathbf{c}_j^t+(1-\alpha)\Delta\mathbf{c}_j^t
\end{equation}
\begin{equation}\label{Eq15}
\Delta\mathbf{c}_j=\dfrac{\sum_{i=1}^b\delta(\hat{y}_i=j)\bm{h}_t^i}{1+\sum_{i=1}^b\delta(\hat{y}_t^i=j)}
\end{equation}
where $\alpha$ is the learning rate of the center. $\bm{c}_j^t$ is the class center of $j$-th class in $t$-th iteration. $\delta(\hat{y}_t^i=j)=1$ if $\bm{x}_t^i$ belongs to $j$-th class, otherwise it should be 0.

\subsection{Full Objective Function}
Based on the aforementioned analysis, to enable effective unsupervised domain adaptation, we propose a holistic approach with an integration of (1) source domain loss minimization, (2) domain alignment with the higher-order moment matching and (3) discriminative clustering in the target domain. The full objective function is as follows,
\begin{equation}\label{eq16}
\mathcal{L}=\mathcal{L}_{s}+\lambda_d\mathcal{L}_{d}+\lambda_{dc}\mathcal{L}_{dc}
\end{equation}
where $\mathcal{L}_s$ is the classification loss in the source domain, $\mathcal{L}_d$ is the domain discrepancy loss measured by the higher-order moment matching, and $\mathcal{L}_{dc}$ denotes the discriminative clustering loss. Note that in order to obtain reliable pseudo-labels for discriminative clustering, we set $\lambda_{dc}=0$ during the initial iterations, and enable the clustering loss $\mathcal{L}_{dc}$ after the total loss tends to be stable.

\section{Experiments}
\subsection{Setup}
\noindent\textbf{Dataset.}
We conduct experiments on three public visual adaptation datasets: digits recognition dataset, Office-31 dataset, and Office-Home dataset. The digits recognition dataset includes four widely used benchmarks: MNIST, USPS, Street View House Numbers (SVHN), and SYN (synthetic digits dataset). We evaluate our proposal across three typical transfer tasks, including: \textbf{SVHN}$\rightarrow$\textbf{MNIST}, \textbf{USPS}$\rightarrow$\textbf{MNIST} and \textbf{SYN}$\rightarrow$\textbf{MNIST}. The details of this dataset can be seen in \cite{chen2019joint}. Office-31 is another commonly used dataset for real-world domain adaptation scenario, which contains 31 categories acquired from the office environment in three distinct image domains: \textbf{A}mazon (product images download from amazon.com), \textbf{W}ebcam (low-resolution images taken by a webcam) and \textbf{D}slr (high-resolution images taken by a digital SLR camera). The office-31 dataset contains 4110 images in total, with 2817 images in \textbf{A} domain, 795 images in \textbf{W} domain and 498 images in \textbf{D} domain. We evaluate our method on all the six transfer tasks as \cite{long2017deep}. The Office-Home dataset \cite{venkateswara2017deep} is a more challenging dataset for domain adaptation, which consists of images from four different domains: Artistic images (\textbf{A}), Clip Art images (\textbf{C}), Product images (\textbf{P}) and Real-world images (\textbf{R}). The dataset contains around 15500 images in total from 65 object categories in office and home scenes.

\noindent\textbf{Baseline Methods.}
We compare our proposal with the following methods, which are most related to our work:
Deep Domain Confusion (\textbf{DDC}) \cite{tzeng2014deep}, Deep Adaptation Network (\textbf{DAN}) \cite{long2015learning}, Deep Correlation Alignment (\textbf{CORAL}) \cite{sun2016deep},  Domain-adversarial Neural Network (\textbf{DANN}) \cite{ganin2016domain}, Adversarial Discriminative Domain Adaptation (\textbf{ADDA}) \cite{tzeng2017adversarial}, Joint Adaptation Network (\textbf{JAN}) \cite{long2017deep}, Central Moment Discrepancy (\textbf{CMD}) \cite{zellinger2017central}  Cycle-consistent Adversarial Domain Adaptation (\textbf{CyCADA}) \cite{hoffman2018cycada}, Joint Discriminative feature Learning and Domain Adaptation (\textbf{JDDA}) \cite{chen2019joint}. Specifically, DDC, DAN, JAN, CORAL and CMD are representative moment matching based methods, while DANN, ADDA and CyCADA are representative adversarial training based methods.

\begin{table}[ht]
\centering
\caption{ Test accuracy (\%) on digits recognition dataset for unsupervised domain adaptation based on modified LeNet}\label{tab:aStrangeTable}
\label{tab1}
\small
\begin{tabular}{ccccc}
\toprule
Method&  SN$\rightarrow$MT&US$\rightarrow$MT&SYN$\rightarrow$MT&Avg\\
\midrule
Source Only &67.3$\pm$0.3&$66.4\pm$0.4&89.7$\pm$0.2&74.5\\
DDC &71.9$\pm$0.4&75.8$\pm$0.3&89.9$\pm$0.2&79.2\\
DAN &79.5$\pm$0.3&$89.8\pm$0.2&75.2$\pm$0.1&81.5\\
DANN &70.6$\pm$0.2&76.6$\pm$0.3&90.2$\pm$0.2&79.1\\
CMD &86.5$\pm$0.3&86.3$\pm$0.4&96.1$\pm$0.2&89.6\\
ADDA &72.3$\pm$0.2&92.1$\pm$0.2&96.3$\pm$0.4&86.9\\
CORAL &89.5$\pm$0.2&96.5$\pm$0.3&96.5$\pm$0.2&94.2\\
CyCADA &92.8$\pm$0.1&97.4$\pm$0.3&97.5$\pm$0.1&95.9\\
JDDA &94.2 $\pm$0.1&96.7$\pm$0.1&97.7$\pm$0.0&96.2\\
\midrule
\textbf{HoMM}(p=3) &96.5$\pm$0.2&97.8$\pm$0.0&97.6$\pm$0.1&97.3\\
\textbf{HoMM}(p=4) &95.7$\pm$0.2&97.6$\pm$0.0&97.6$\pm$0.0&96.9\\
\textbf{KHoMM}(p=3) &97.2$\pm$0.1&97.9$\pm$0.1&98.2$\pm$0.1&97.8\\
\textbf{Full} &\textbf{98.8}$\pm$0.1&\textbf{99.0}$\pm$0.1&\textbf{99.0}$\pm$0.0&\textbf{98.9}\\
\midrule
\textbf{KHoMM+$\mathcal{L}_{ent}$} &\textbf{99.0}$\pm$0.0&\textbf{99.1}$\pm$0.1&\textbf{99.2}$\pm$0.0&\textbf{99.1}\\
\bottomrule
\end{tabular}
\footnotesize \small We denote SVHN, MNIST, USPS as SN, MT and US.
\end{table}

\begin{table*}[ht]
\centering
\caption{Test accuracy (\%) on Office-31 dataset for unsupervised domain adaptation based on ResNet-50}\label{tab:aStrangeTable}
\label{tab2}
\begin{tabular}{cccccccc}
\toprule
Method& A$\rightarrow$W&D$\rightarrow$W&W$\rightarrow$D&A$\rightarrow$D&D$\rightarrow$A&W$\rightarrow$A&Avg\\
\midrule
Source Only &73.1$\pm$0.2&93.2$\pm$0.2&$98.8\pm$0.1&72.6$\pm$0.2&55.8$\pm$0.1&56.4$\pm$0.3&75.0\\
DDC \cite{tzeng2014deep}&74.4$\pm$0.3&94.0$\pm$0.1&98.2$\pm$0.1&74.6$\pm$0.4&56.4$\pm$0.1&56.9$\pm$0.1&75.8\\
DAN \cite{long2015learning}&78.3$\pm$0.3&$95.2\pm$0.2&$99.0\pm$0.1&75.2$\pm$0.2&58.9$\pm$0.2&64.2$\pm$0.3&78.5\\
DANN \cite{ganin2016domain}&73.6$\pm$0.3&94.5$\pm$0.1&99.5$\pm$0.1&74.4$\pm$0.5&57.2$\pm$0.1&60.8$\pm$0.2&76.7\\
CORAL \cite{sun2016deep}&79.3$\pm$0.3&94.3$\pm$0.2&99.4$\pm$0.2&74.8$\pm$0.1&56.4$\pm$0.2&63.4$\pm$0.2&78.0\\
JAN \cite{long2017deep}&85.4$\pm$0.3&97.4$\pm$0.2&99.8$\pm$0.2&84.7$\pm$0.4&68.6$\pm$0.3&70.0$\pm$0.4&84.3\\
CMD \cite{zellinger2017central}&76.9$\pm$0.4&94.6$\pm$0.3&99.2$\pm$0.2&75.4$\pm$0.4&56.8$\pm$0.1&61.9$\pm$0.2&77.5\\
CyCADA \cite{hoffman2018cycada}&82.2$\pm$0.3&94.6$\pm$0.2&99.7$\pm$0.1&78.7$\pm$0.1&60.5$\pm$0.2&67.8$\pm$0.2&80.6\\
JDDA\cite{chen2019joint}&82.6$\pm$0.4&95.2$\pm$0.2&99.7$\pm$0.0&79.8$\pm$0.1&57.4$\pm$0.0&66.7$\pm$0.2&80.2\\
\midrule
\textbf{HoMM}(p=3) &87.6$\pm$0.2&96.3$\pm$0.1&99.8$\pm$0.0&83.9$\pm$0.2&66.5$\pm$0.1&68.5$\pm$0.3&83.7\\
\textbf{HoMM}(p=4) &89.8$\pm$0.3&97.1$\pm$0.1&100.0$\pm$0.0&86.6$\pm$0.1&69.6$\pm$0.3&69.7$\pm$0.3&85.5\\
\textbf{KHoMM}(p=4) &90.5$\pm$0.2&98.3$\pm$0.1&100.0$\pm$0.0&87.7$\pm$0.2&70.4$\pm$0.2&70.3$\pm$0.2&86.2\\
\textbf{Full} &\textbf{91.7}$\pm$0.3&\textbf{98.8}$\pm$0.0&\textbf{100.0}$\pm$0.0&\textbf{89.1}$\pm$0.3&\textbf{71.2}$\pm$0.2&\textbf{70.6}$\pm$0.3&\textbf{86.9}\\
\midrule
\textbf{KHoMM+$\mathcal{L}_{ent}$} &90.8$\pm$0.1&\textbf{99.3}$\pm$0.1&\textbf{100.0}$\pm$0.0&87.9$\pm$0.2&69.3$\pm$0.3&69.5$\pm$0.4& 86.1\\
\bottomrule
\end{tabular}
\end{table*}

\begin{table}[ht]
\centering
\caption{Test accuracy (\%) on Office-Home dataset for unsupervised domain adaptation based on ResNet-50}\label{tab:aStrangeTable}
\label{tab3}\resizebox{\linewidth}{!}{
\begin{tabular}{ccccccc}
\toprule
Method& A$\rightarrow$P&A$\rightarrow$R&C$\rightarrow$R&P$\rightarrow$R&R$\rightarrow$P\\
\midrule
Source Only  &50.0 &58.0 &46.2 &60.4 &59.5 \\
DDC          &54.9 &61.3 &50.5 &64.1 &65.9 \\
DAN          &57.0 &67.9 &60.4 &67.7 &74.3 \\
DANN         &59.3 &70.1 &60.9 &68.5 &76.8 \\
CORAL        &58.6 &65.4 &59.8 &68.3 &74.7 \\
JAN          &61.2 &68.9 &61.0 &70.3 &76.8 \\
\midrule
\textbf{HoMM}(p=3)         &60.7 &68.3 &61.4 &69.2 &76.7 \\
\textbf{HoMM}(p=4)         &63.5 &70.2 &64.6 &72.6 &79.3 \\
\textbf{KHoMM}(p=4)        &63.9 &70.5 &65.3 &73.3 &79.8 \\
\textbf{Full}    &\textbf{64.7} &\textbf{71.8} &\textbf{66.1} &\textbf{74.5} &\textbf{81.2} \\
\midrule
\textbf{KHoMM+$\mathcal{L}_{ent}$}    &64.2 &70.1 &65.5 &73.2 &80.1 \\
\bottomrule
\end{tabular}}
\end{table}

\noindent\textbf{Implementation Details.}
In our experiments on digits recognition dataset, we utilize the modified LeNet whereby a bottleneck layer with $90$ hidden neurons is added before the output layer. Since the image size is different across different domains, we resize all the images to $32\times32$ and convert the RGB images to grayscale. For the experiments on Office-31, we use ResNet-50 pretrained on ImageNet as our backbone networks. And we add a bottleneck layer with 180 hidden nodes before the output layer for domain matching. It is worth noting that the \textbf{relu} activation function can not be applied to the adapted layer, as relu activation function will make most of the values in the high-level tensor $\bm{h}_{s}^i{^{\otimes p}}$ to be zero, which will make our HoMM fail. Therefore, we adopt \textbf{tanh} activation function in the adapted layer. Due to the small samples size of Office-31 and Office-Home datasets, we only update the weights of the full-connected layers (fc) as well as the final block (scale5/block3), and fix other parameters pretrained on ImageNet. Follow the standard protocol of \cite{long2017deep}, we use all the labeled source domain samples and all the unlabeled target domain samples for training. All the comparison methods are based on the same CNN architecture for a fair comparison. For DDC, DAN, CORAL and CMD, we embed the official implementation code into our model and carefully select the trade-off parameters to get the best results. When training with ADDA, our adversarial discriminator consists of 3 fully connected layers: two layers with 500 hidden units followed by the final discriminator output. For other compared methods, we report the results in the original paper directly.

\noindent\textbf{Parameters.} Our model is trained with Adam Optimizer based on Tensorflow. Regarding the optimal hyper-parameters, they are determined by applying multiple experiments using grid search strategy. The optimal hyper-parameters may be distinct across different transfer tasks. Specifically, the trade-off parameters are selected from $\lambda_{d}=\{1,10,10^2,\cdots,10^8\}$, $\lambda_{dc}\in\{0.01,0.03,0.1,0.3,1.0\}$. For the digits recognition tasks, the hyper-parameter $\lambda_d$ is set to $10^4$ for third-order HoMM and set to $10^7$ for fourth-order HoMM \footnote{Note that the trade-off on the fourth-order HoMM is much larger than the third-order HoMM. This is because most deep features of the digits are very small, higher-order moments calculating the cumulative multiplication between different features become very close to zeros. Therefore, on digits dataset, the higher the order, the larger the trade-off should be.}. For the experiments on Office-31 and Office-Home, $\lambda_d$ is set to $300$ for the third-order HoMM and set to $3000$ for the fourth-order HoMM.  Besides, the hyper-parameter $\gamma$ in RBF kernel is set to 1e-4 across the experiments, the learning rate of the centers is set to $\alpha=0.5$ for digits dataset and set to $\alpha=0.3$ for Office-31 and Office-Home dataset. The threshold $\eta$ of the predicted probability is chosen from $\{0.6,0.65,0.7,0.75,0.8,0.85,0.9,0.95\}$, and the best results are reported. The parameter sensitivity can be seen in Fig. \ref{fig4}.

\subsection{Experimental results}
\textbf{Digits Dataset} For the experiments on digits recognition dataset, we set the batch size as 128 for each domain and set the learning rate as 1e-4 throughout the experiments. Table \ref{tab1} shows the adaptation performance on three typical transfer tasks based on the modified LeNet. As can be seen, our proposed HoMM yields notable improvement over the comparison methods on all of the transfer tasks. In particular, our method improves the adaption performance significantly in the hard transfer tasks SVHN$\rightarrow$MNIST. Without bells and whistles, the proposed third-order KHoMM achieve 97.2\% accuracy, improving the second-order moment matching (CORAL) by +8\%. Besides, the results also indicate that the third-order HoMM outperforms the fourth-order HoMM and slightly underperforms the KHoMM.

\noindent\textbf{Office-31} Table \ref{tab2} lists the test accuracies on Office-31 dataset. We set the batchsize as 70 for each domain. The learning rate of the fc layer parameters is set as 3e-4 and the learning rate of the conv layer (scale5/block3) parameters is set as 3e-5. As we can see, the fourth-order HoMM outperforms the third-order HoMM and achieves the best results among all the moment-matching based methods. Besides, it is worth noting that the fourth-order HoMM outperforms the second-order statistics matching (CORAL) by more than 10\% on several representative transfer tasks A$\rightarrow$W, A$\rightarrow$D and D$\rightarrow$A, which demonstrates the merits of our proposed higher-order moment matching.

\noindent\textbf{Office-Home} Table 3 gives the results on the challenged Office-Home dataset. The parameter settings are the same as in Office-31. We only evaluate our method on 5 out of 12 representative transfer tasks due to the space limitation. As we can see, on all the five transfer tasks, the HoMM outperforms the DAN, CORAL, DANN by a large margin and also outperforms the JAN by 3\%-5\%. Note that the experimental results of the compared methods are reported from \cite{wang2019transferable} directly.

The results in Table \ref{tab1}, Table \ref{tab2} and Table \ref{tab3} reveal several interesting observations: (1) All the domain adaptation methods outperform the source only model by a large margin, which demonstrates that minimizing the domain discrepancy contributes to learning more transferable representations. (2) Our proposed HoMM significantly outperforms the discrepancy-based methods (DDC, CORAL, CMD), and the adversarial training based methods (DANN, ADDA and CyCADA), which reveals the advantages of matching the higher-order statistics for domain adaptation. (3) The JAN performs slightly better than the third-order HoMM on several transfer tasks, but it's always not as good as the fourth-order HoMM in spite of aligning the joint distributions of multiple domain-specific layers across domains. The performance of our HoMM will be improved as well if we utilize such a strategy. (4) The kernelized HoMM (KHoMM) consistently outperforms the plain HoMM, but the improvement seems limited. We believe the reason is that, the higher-order statistics are originally the high-dimensional features, which conceals the advantages of embedding the features into RKHS. (5) In all transfer tasks, the performance increases consistently by employing the discriminative clustering in target domain. In contrast, entropy regularization improves the transfer performance when the test accuracy is high, but it helps little or even downgrades the performance when the test accuracy is not that confident.

\begin{figure*}[!ht]
    \centering
  \begin{subfigure}[b]{0.19\textwidth}
    \includegraphics[width=\textwidth]{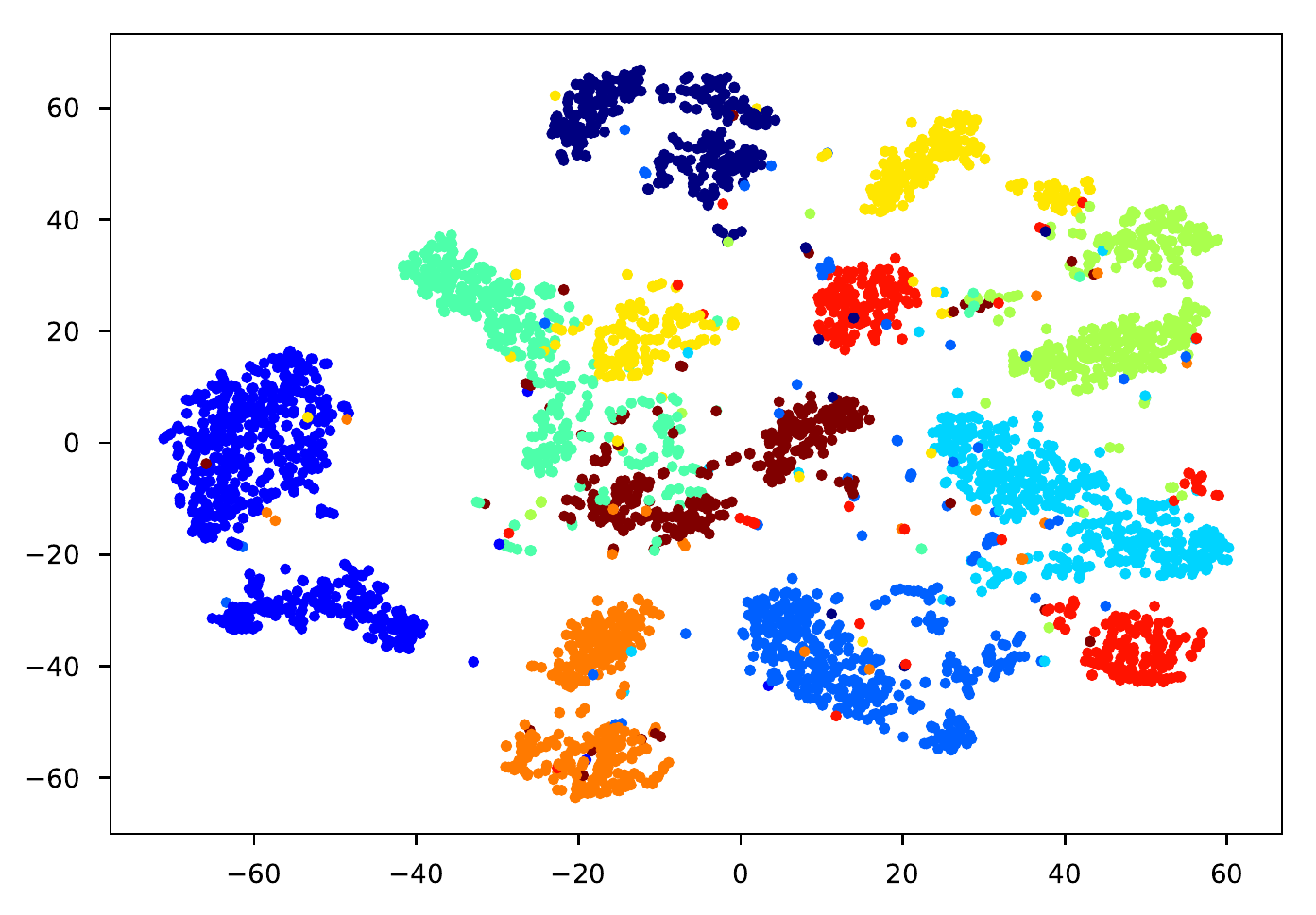}
    \caption{Source Only}
    \label{2D1}
  \end{subfigure}
   \begin{subfigure}[b]{0.19\textwidth}
    \includegraphics[width=\textwidth]{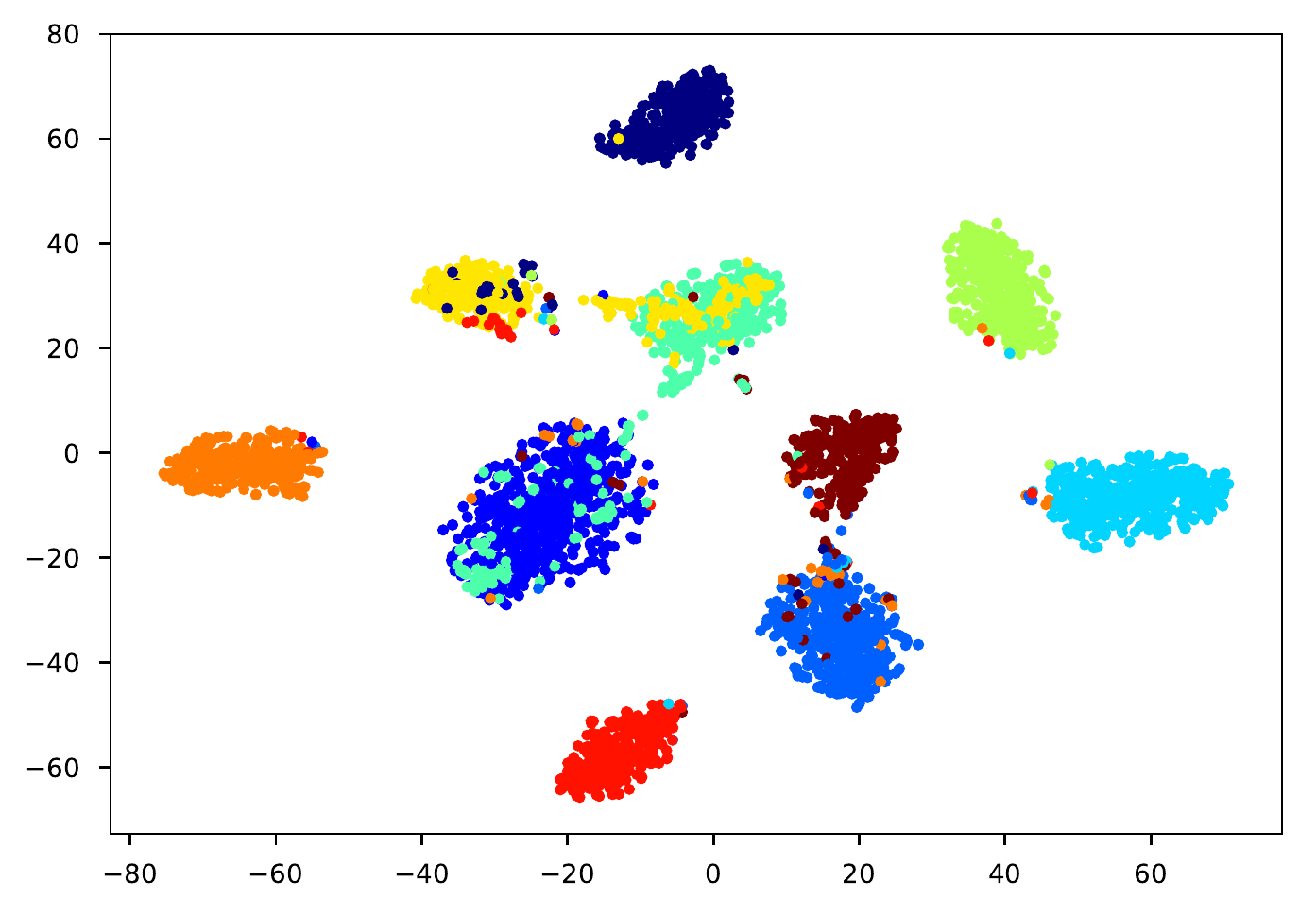}
    \caption{KMMD}
    \label{2D2}
  \end{subfigure}
    \begin{subfigure}[b]{0.19\textwidth}
    \includegraphics[width=\textwidth]{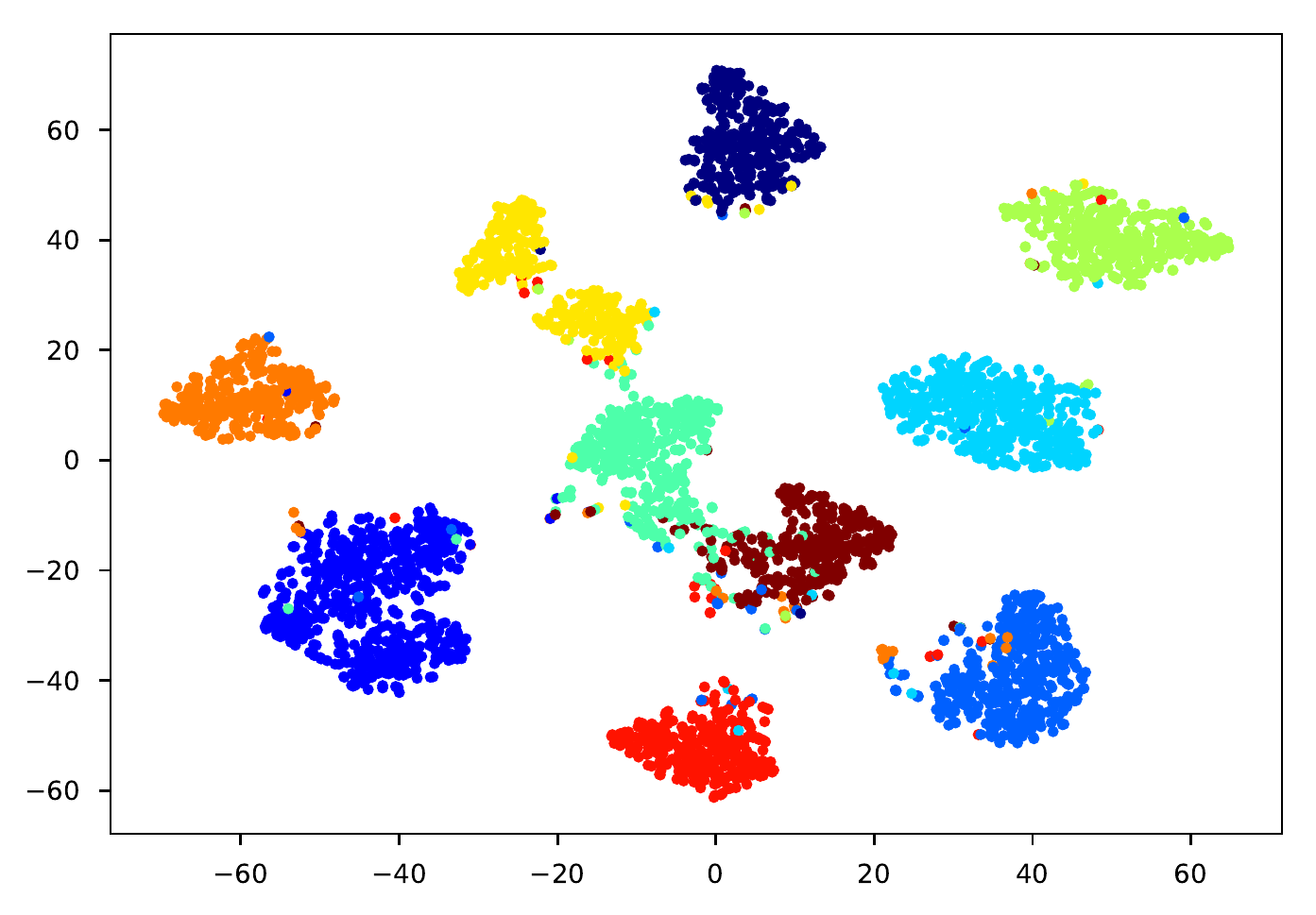}
    \caption{CORAL}
    \label{2D3}
  \end{subfigure}
    \begin{subfigure}[b]{0.19\textwidth}
    \includegraphics[width=\textwidth]{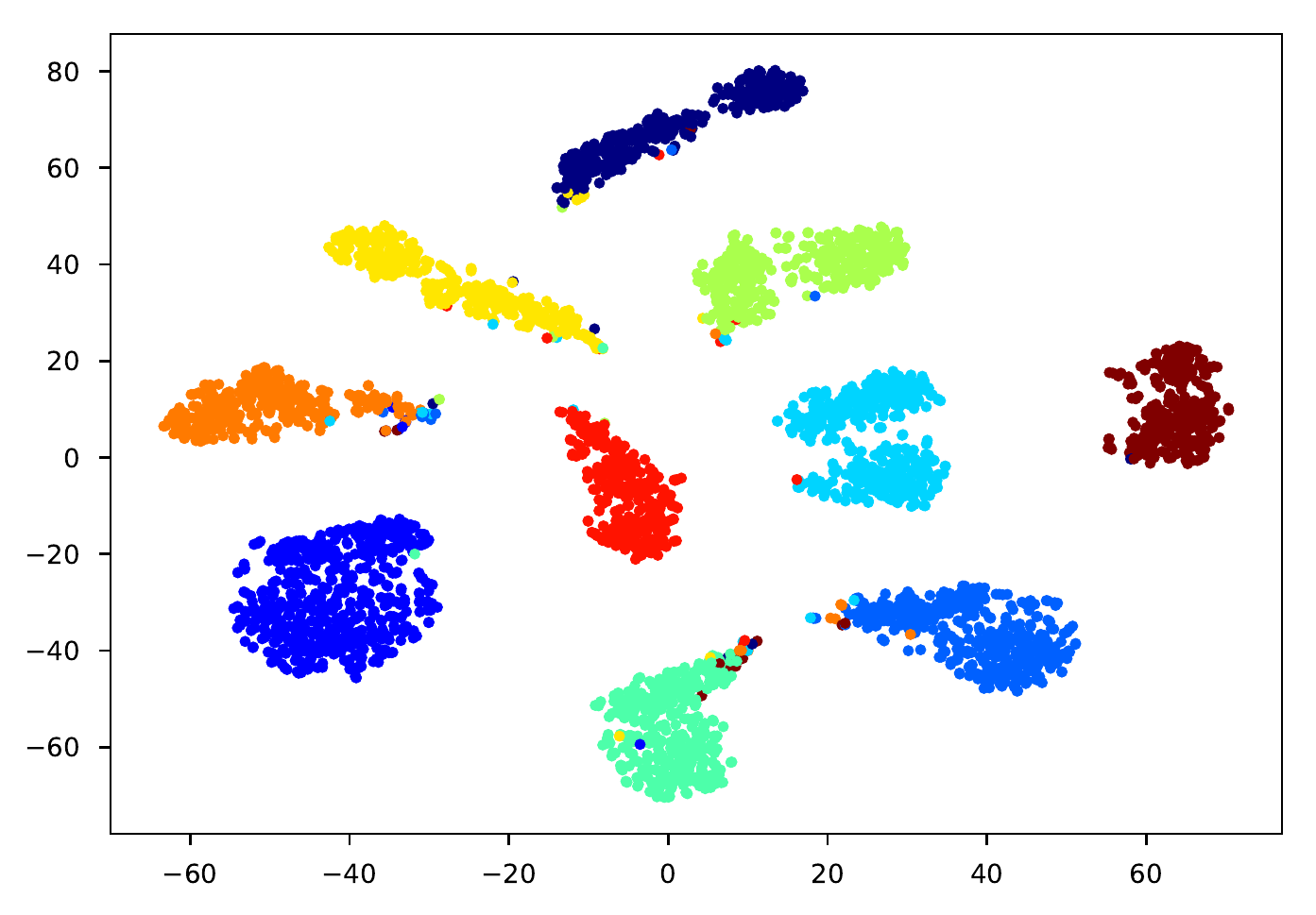}
    \caption{HoMM(p=3)}
    \label{2D4}
  \end{subfigure}
    \begin{subfigure}[b]{0.19\textwidth}
    \includegraphics[width=\textwidth]{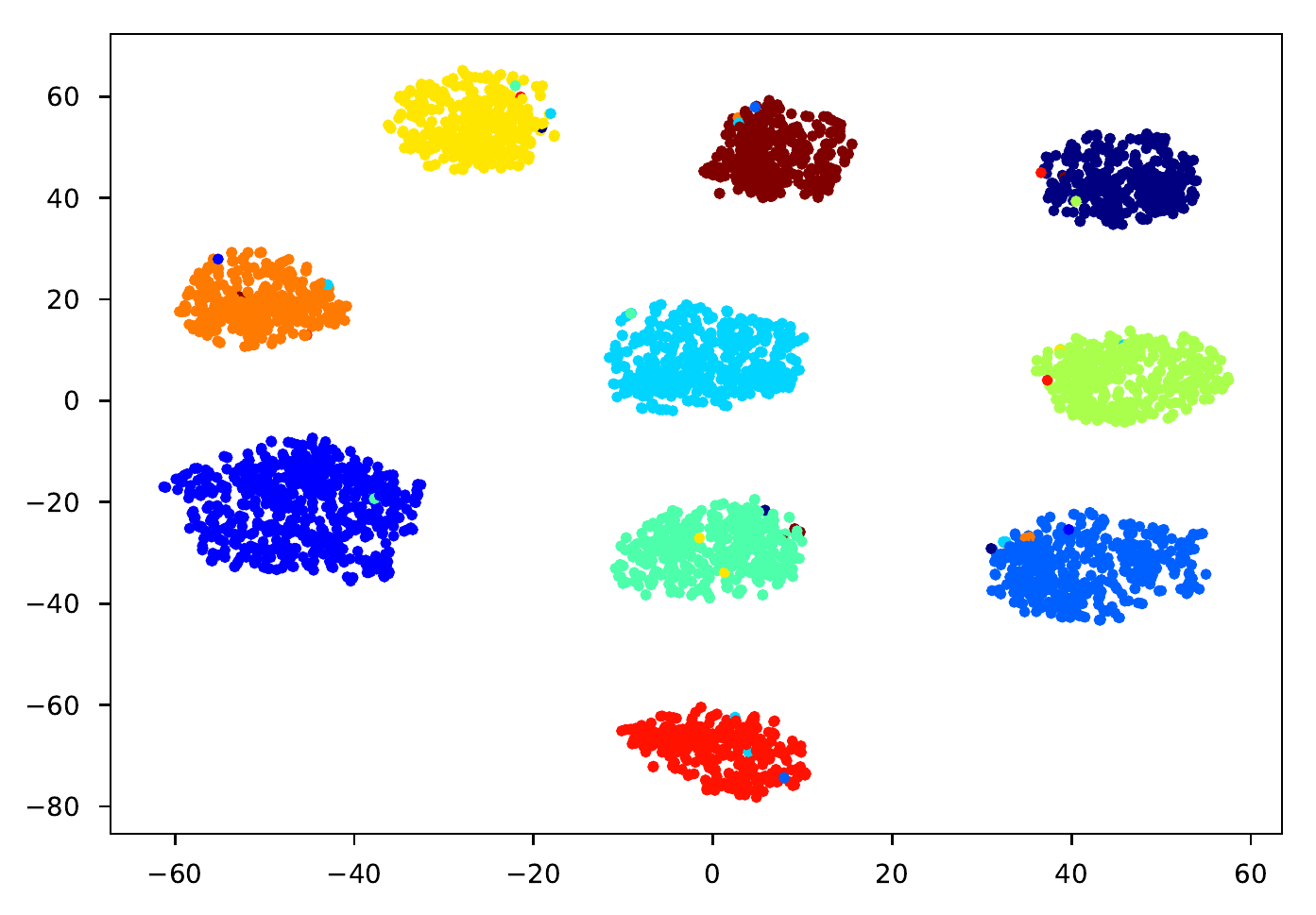}
    \caption{Full Loss}
    \label{2D5}
  \end{subfigure}

    \begin{subfigure}[b]{0.19\textwidth}
    \includegraphics[width=\textwidth]{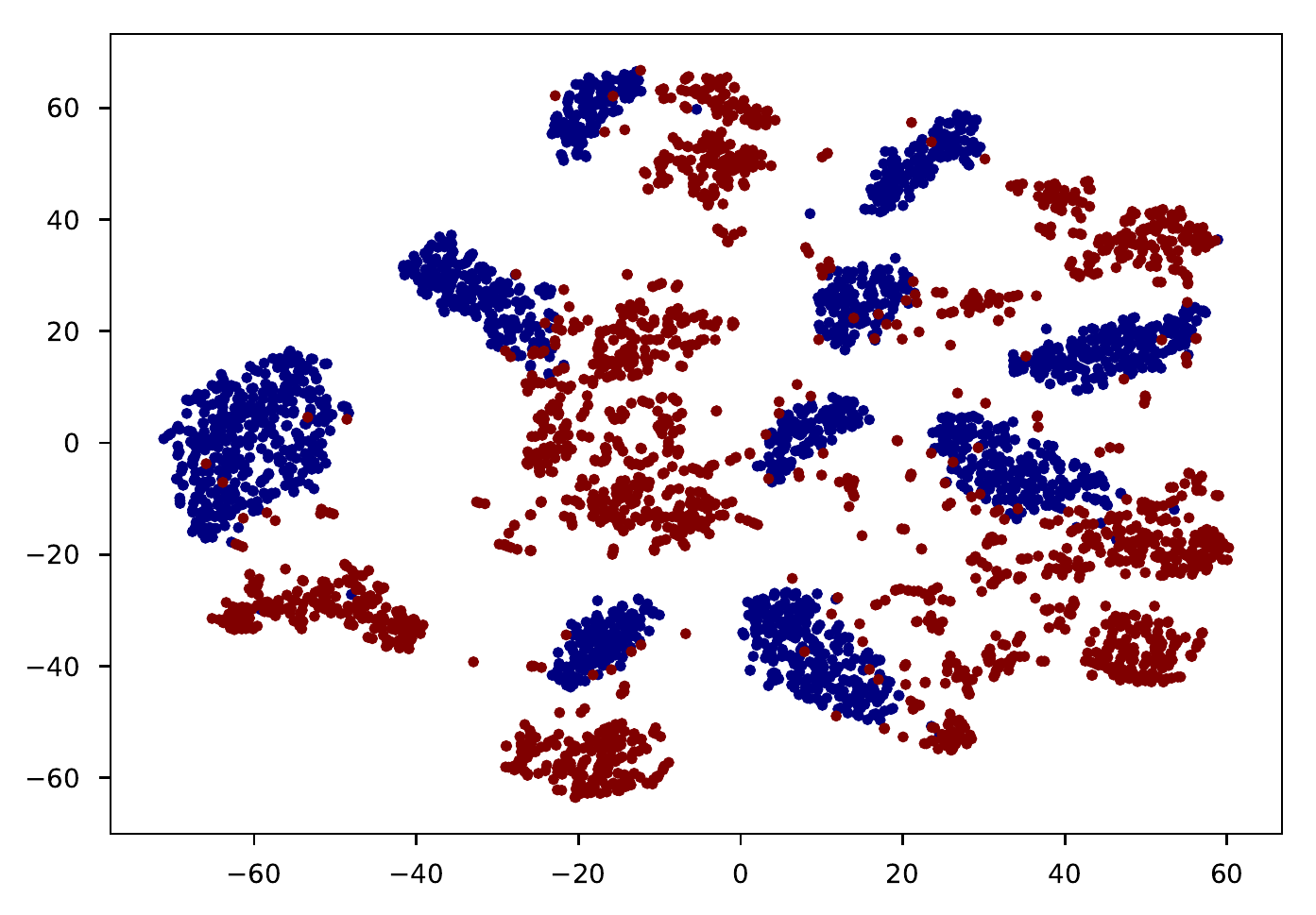}
    \caption{Source Only}
    \label{2D21}
  \end{subfigure}
   \begin{subfigure}[b]{0.19\textwidth}
    \includegraphics[width=\textwidth]{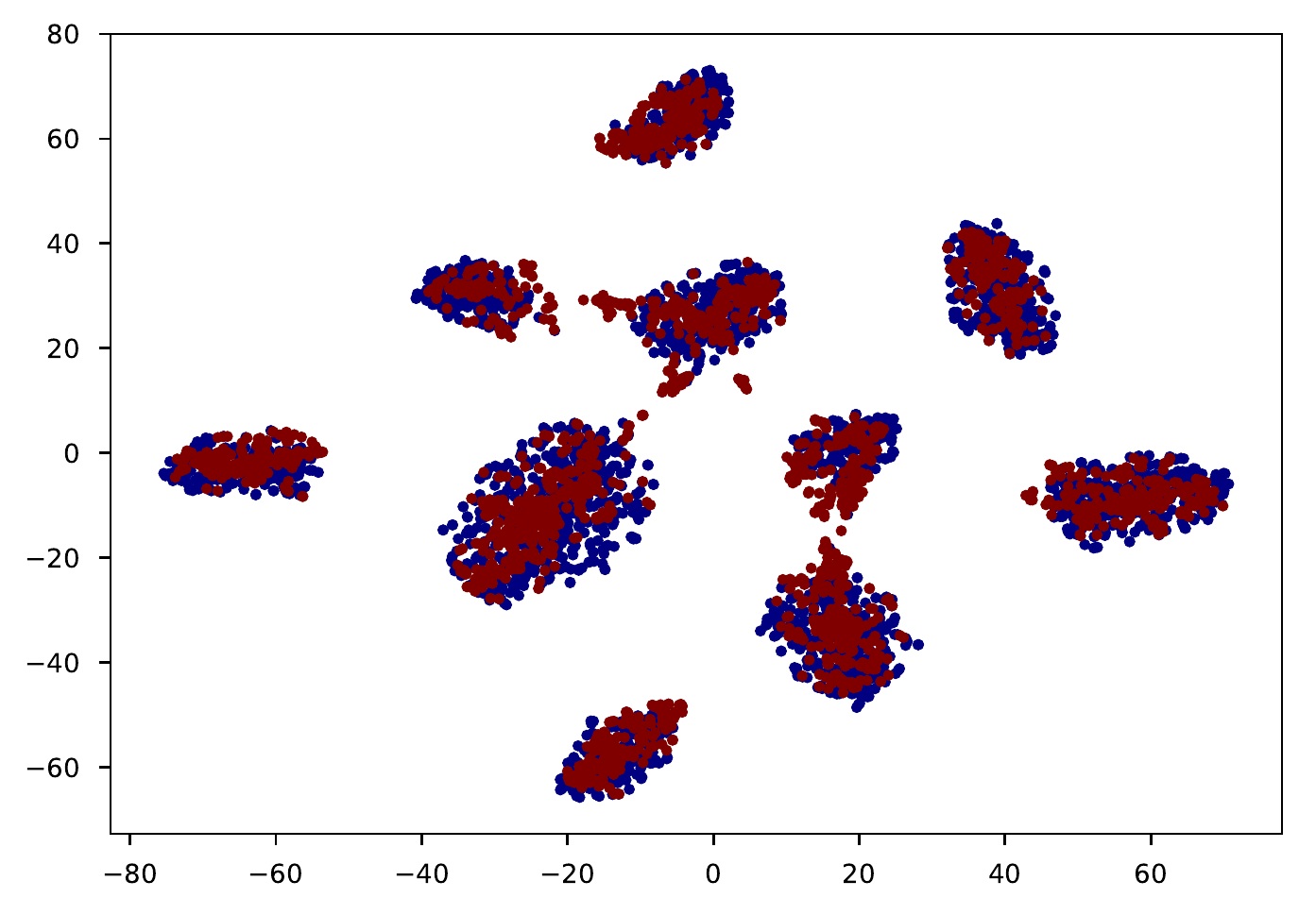}
    \caption{KMMD}
    \label{2D22}
  \end{subfigure}
    \begin{subfigure}[b]{0.19\textwidth}
    \includegraphics[width=\textwidth]{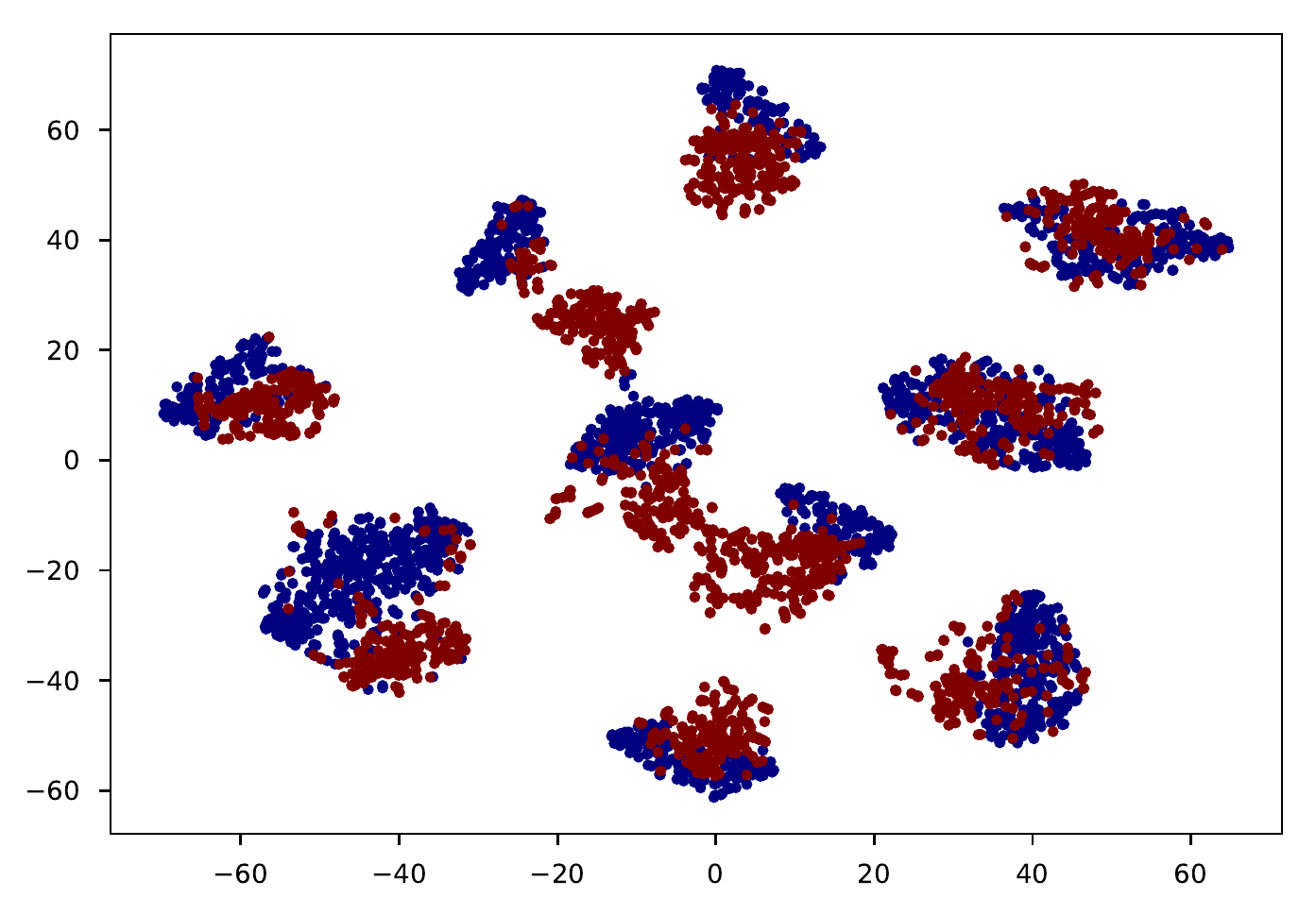}
    \caption{CORAL}
    \label{2D23}
  \end{subfigure}
    \begin{subfigure}[b]{0.19\textwidth}
    \includegraphics[width=\textwidth]{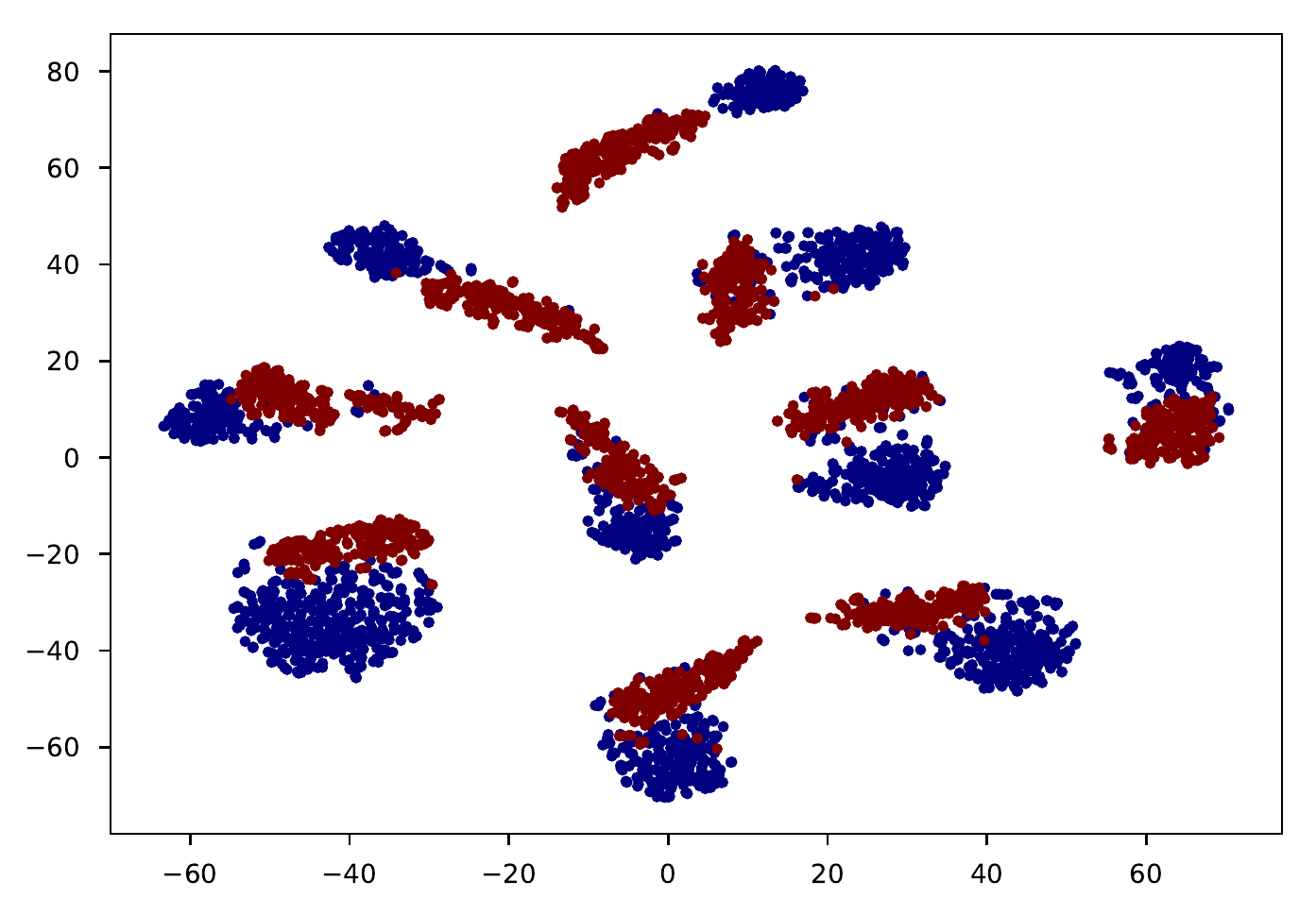}
    \caption{HoMM(p=3)}
    \label{2D24}
  \end{subfigure}
    \begin{subfigure}[b]{0.19\textwidth}
    \includegraphics[width=\textwidth]{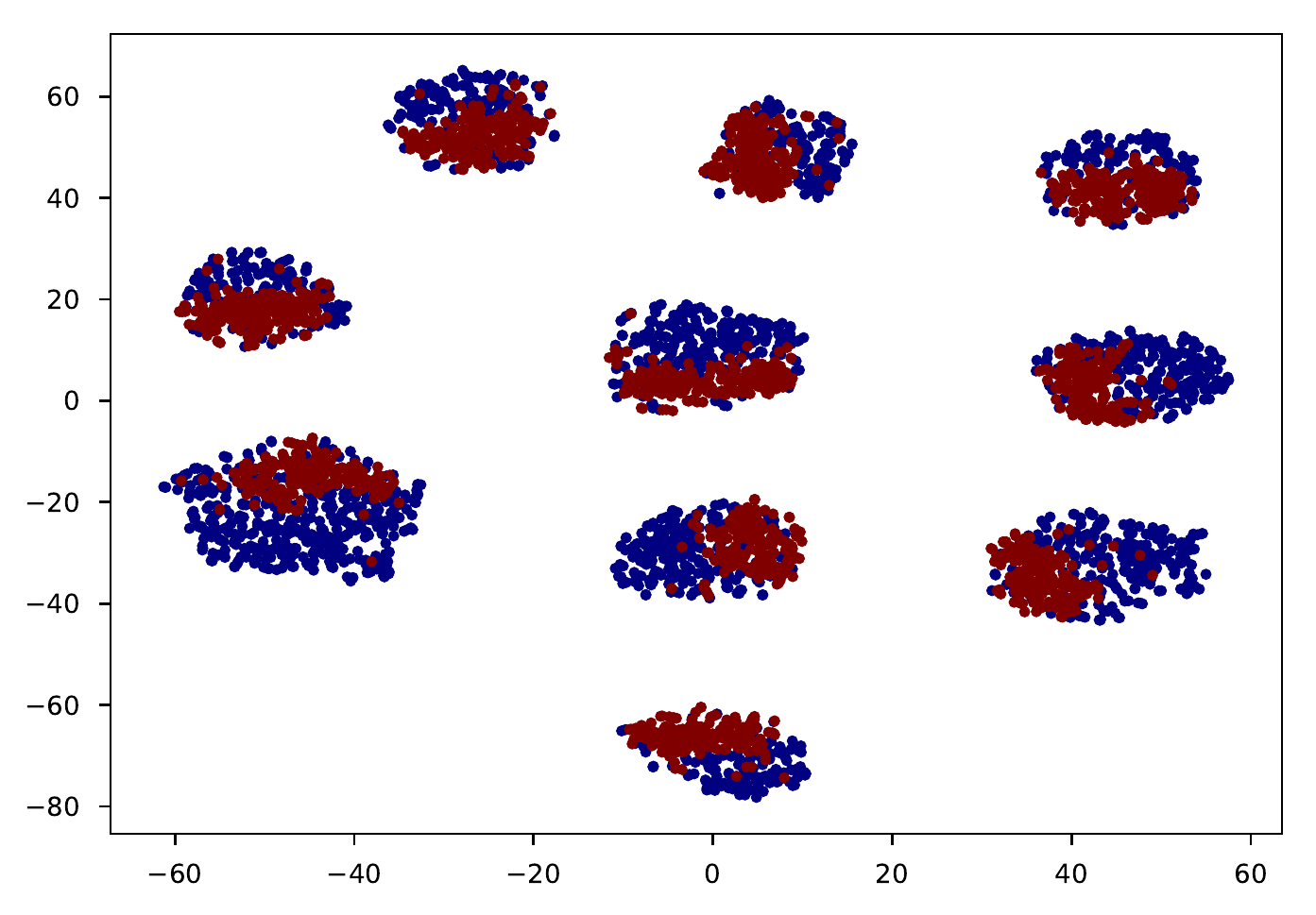}
    \caption{Full Loss}
    \label{2D25}
  \end{subfigure}
\caption{ 2D visualization of the deep features generated by different model on SVHN$\rightarrow$MNIST. The first row illustrates the t-SNE embedding of deep features
which are marked by category information, each color represents a category. The second row illustrates the t-SNE embedding of deep features which are marked by domain information, red and blue points represent the samples drawn from the source and target domains.}
\label{fig3}
\end{figure*}

\begin{figure*}[!ht]
    \centering
  \begin{subfigure}[b]{0.19\textwidth}
    \includegraphics[width=\textwidth]{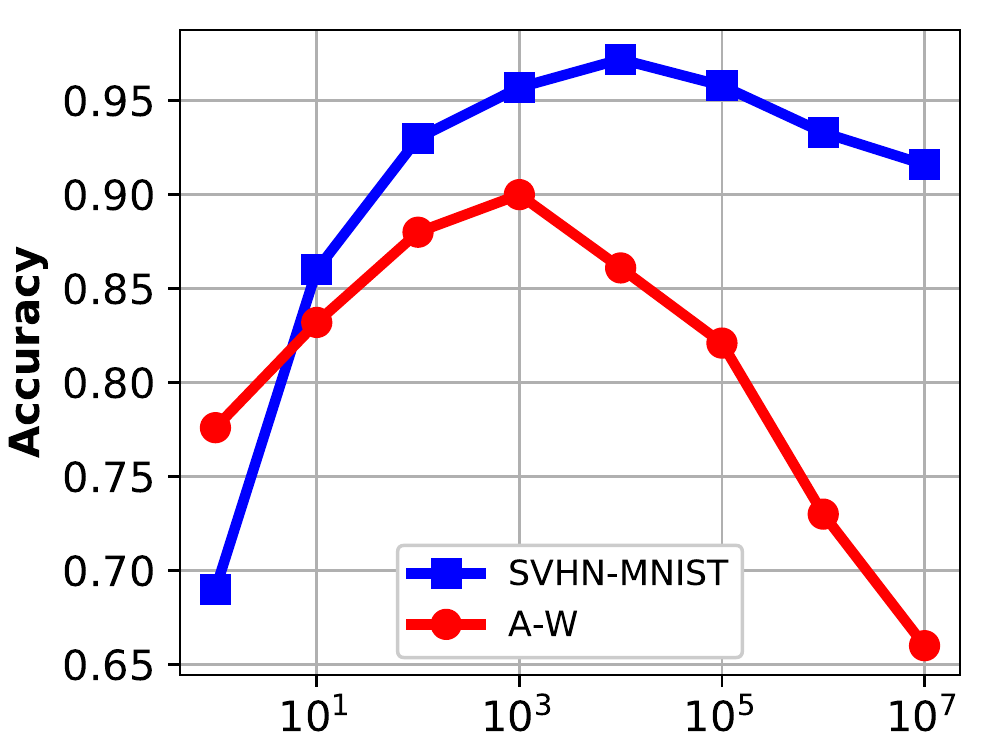}
    \caption{$\lambda_d$}
    \label{2D11}
  \end{subfigure}
   \begin{subfigure}[b]{0.19\textwidth}
    \includegraphics[width=\textwidth]{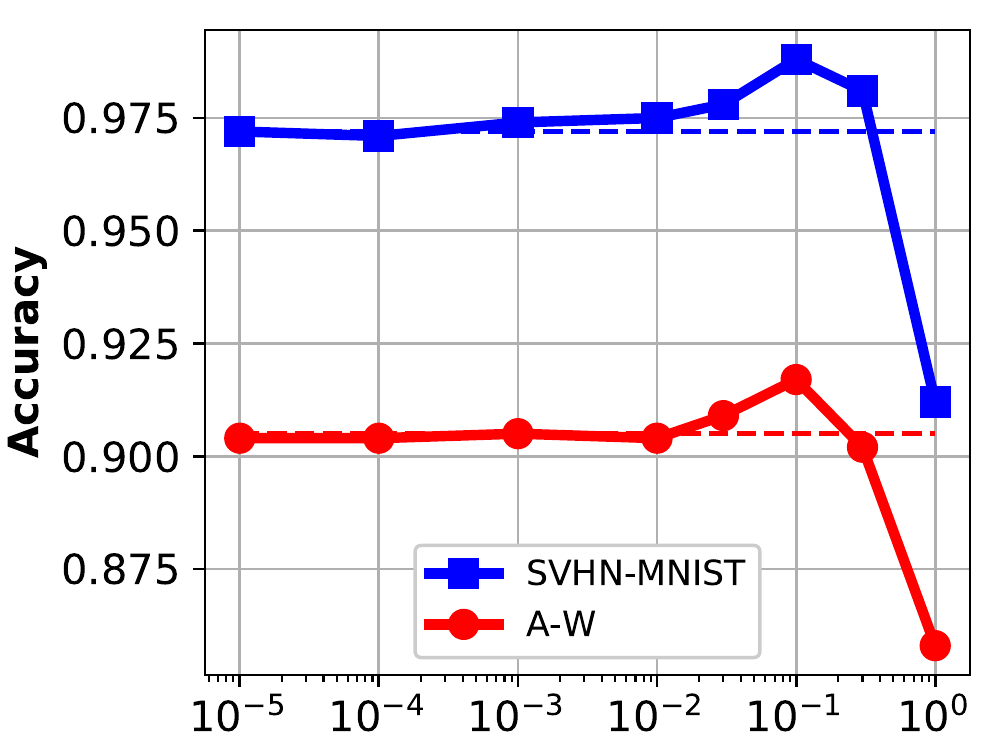}
    \caption{$\lambda_{dc}$}
    \label{2D22}
  \end{subfigure}
    \begin{subfigure}[b]{0.19\textwidth}
    \includegraphics[width=\textwidth]{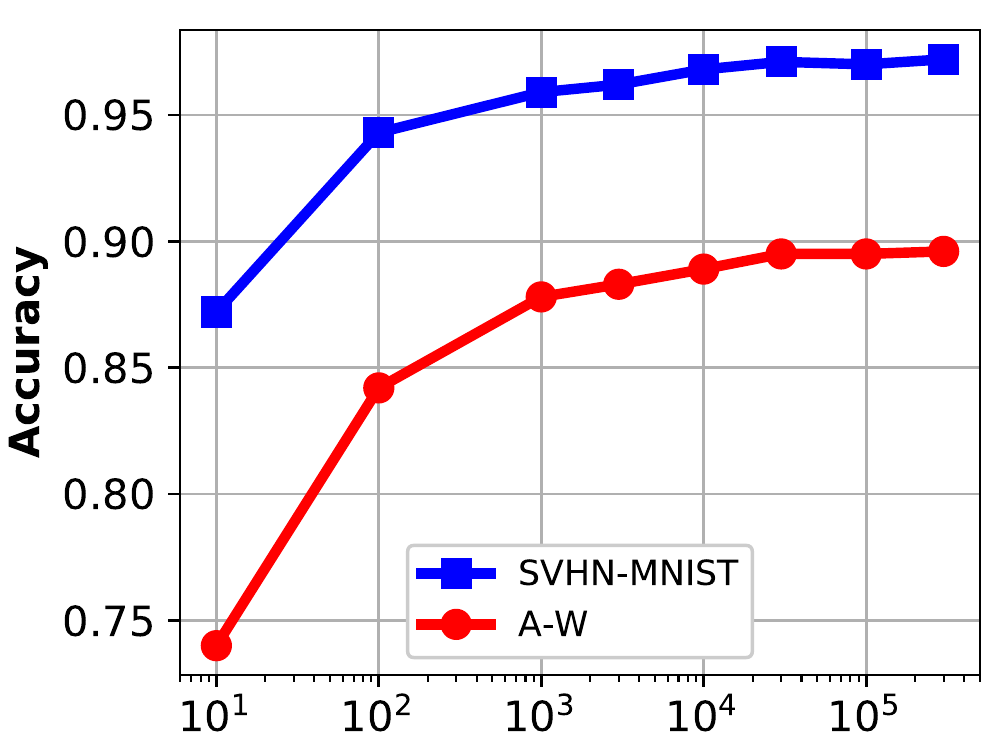}
    \caption{$N$}
    \label{2D33}
  \end{subfigure}
    \begin{subfigure}[b]{0.19\textwidth}
    \includegraphics[width=\textwidth]{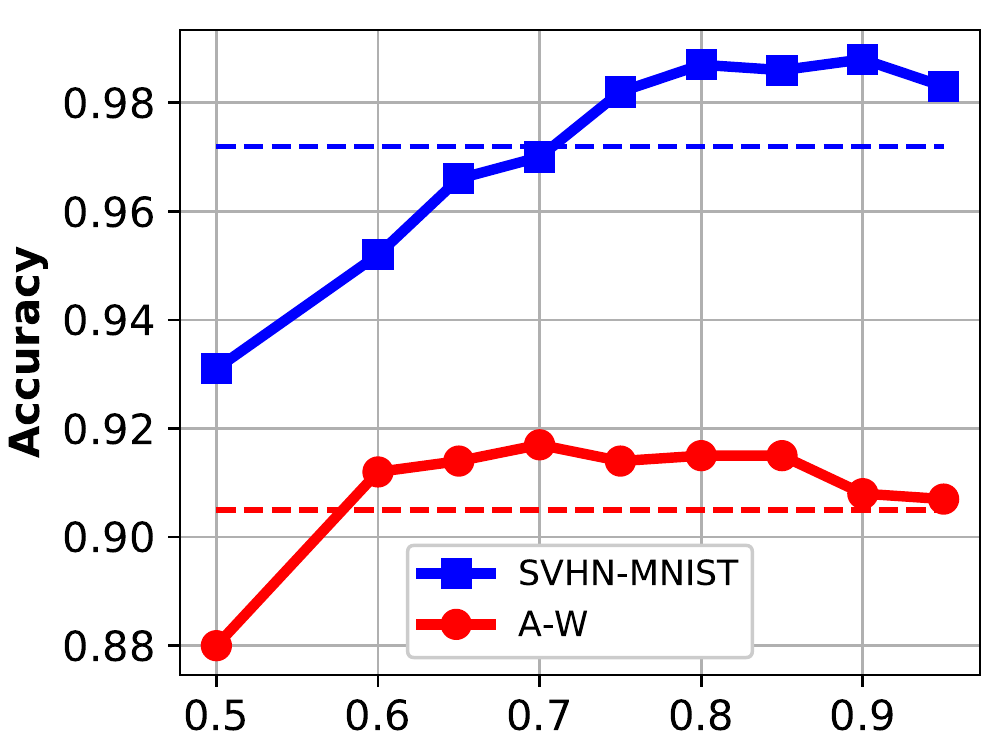}
    \caption{$\eta$}
    \label{2D55}
  \end{subfigure}
  \begin{subfigure}[b]{0.19\textwidth}
    \includegraphics[width=\textwidth]{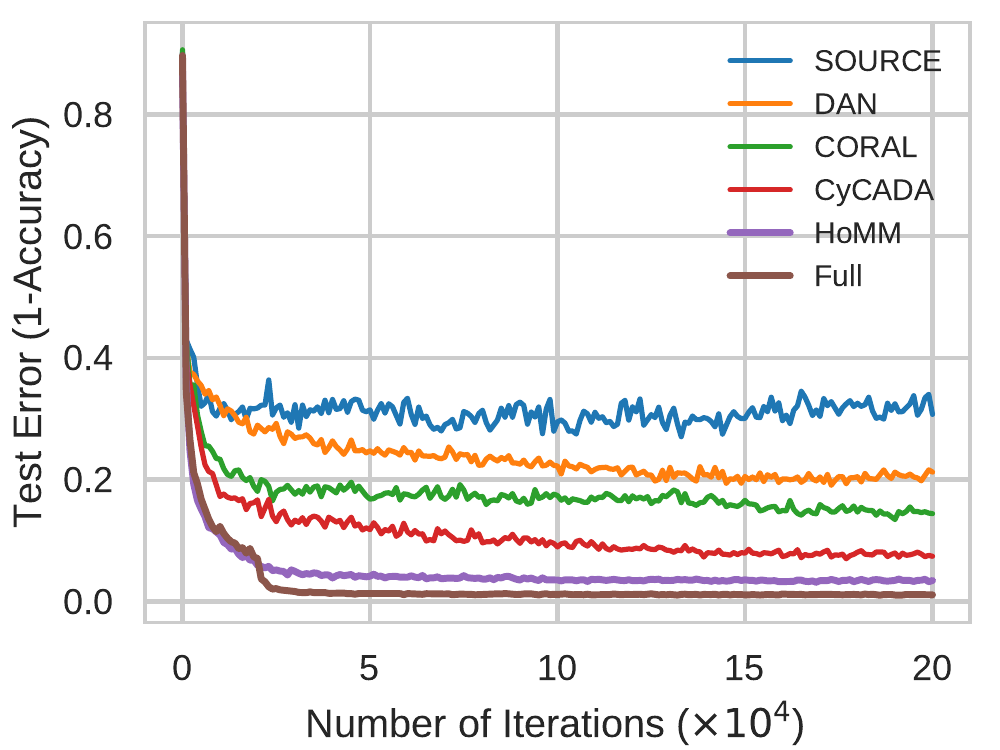}
    \caption{Convergence}
    \label{2D55}
  \end{subfigure}
\caption{ Analysis of parameter sensitivity (a)-(d) and convergence analysis (e). The dash line in (b) and (d) indicate the performance of HoMM without the clustering loss $\mathcal{L}_{dc}$}
\label{fig4}
\end{figure*}

\subsection{Analysis}
\textbf{Feature Visualization}
We utilize t-SNE to visualize the deep features on the tasks SVHN$\rightarrow$MNIST by ResNet-50, KMMD, CORAL, HoMM(p=3) and the Full Loss model. As shown in Fig. \ref{fig3}, the feature distributions of the source only model in (a) suggests that the domain shift between SVNH and MNIST is significant, which demonstrates the necessity of performing domain adaptation. Besides, the global distributions of the source and target samples are well aligned with the KMMD (b) and CORAL (c), but there are still many samples being misclassified. With our proposed HoMM, the source and target samples are aligned better and categories are discriminated better as well.

\noindent\textbf{First/Second-order versus Higher-order}
Since our proposed HoMM can perform arbitrary-order moment matching, we compare the performance of different order moment matching on three typical transfer tasks. As shown in table \ref{tab4}, the order is chosen from $p\in\{1,2,3,4,5,6,10\}$. The results show that the third-order and fourth-order moment matching significantly outperform the other order moment matching. When $p\leq3$, the higher the order, the higher the accuracy. When $p\geq4$, the accuracy will decrease as the order increases. Besides, the fifth-order moment matching also achieves very competitive results. Regarding why the fifth-order and above perform worse than the fourth-order, one reason we believe is that the fifth-order and above moments can’t be accurately estimated due to the small sample size problem \cite{raudys1991small}.

\begin{table}[ht]
\centering
\caption{Test accuracy (\%) comparison of different-order moment matching on three transfer tasks}
\label{tab4}
\small
\begin{tabular}{cccccccc}
\toprule
order &1 &2 &3 &4 &5 &6 &10\\
\midrule
SN$\rightarrow$MT &71.9 &89.5 &\textbf{96.5} &95.7 &94.8 &91.5 &58.6 \\
A$\rightarrow$W &74.4 &79.3 &87.6 &\textbf{89.8} &86.6 &85.3 &80.2 \\
A$\rightarrow$P &54.9 &58.6 &60.7 &\textbf{63.5} &60.9 &58.2 &57.3\\
\bottomrule
\end{tabular}
\footnotesize \small We denote SVHN and MNIST as SN and MT respectively.
\end{table}

\noindent\textbf{Parameter Sensitivity and Convergence}
We conduct empirical parameter sensitivity on SVHN$\rightarrow$MNIST and A$\rightarrow$W in Fig. \ref{fig4}(a)-(d). The evaluated parameters include two trade-off parameters $\lambda_c$, $\lambda_{dc}$, the number of selected values in Random Sampling Matching $N$, and the threshold $\eta$ of the predicted probability. As we can see, our model is quite sensitive to the change of $\lambda_{dc}$ and the bellshaped curve illustrates the regularization effect of $\lambda_d$ and $\lambda_{dc}$. The convergence performance is provided in Fig. \ref{fig4}(e), which shows that our proposal converges fastest compared with other methods. It is worth noting that, the test error of the Full Loss model has a obvious mutation at the $2.0\times10^4$ iteration where we enable the clustering loss $\mathcal{L}_{dc}$, which also demonstrates the effectiveness of the proposed discriminative clustering loss.

\section{Conclusion}
Minimizing statistic distance between source and target distributions is an important line of work for domain adaptation. Unlike previous methods that utilize the second-order or lower statistics for domain alignment, this paper exploits the higher-order statistics for domain alignment. Specifically, a higher-order moment matching method has been presented, which integrates the MMD and CORAL into a unified framework and generalizes the existing first-order and second-order moment matching to arbitrary-order moment matching. We experimentally demonstrate that the third-order and fourth-order moment matching significantly outperform the existing moment matching methods. Besides, we also extend the HoMM into RKHS and learn the discriminative clusters in the target domain, which further improves the adaptation performance. The proposed HoMM can be easily integrated into other domain adaptation model, and it is also expected to benefit the knowledge distillation and image style transfer.

{\small
\bibliographystyle{ieee_fullname}
\bibliography{reference}
}

\end{document}